\documentclass[journal,twoside]{IEEEtran}

\IEEEoverridecommandlockouts
\usepackage{amsmath,amssymb,amsfonts}
\usepackage{color} 
\usepackage{graphicx} 
\usepackage[utf8]{inputenc}
\usepackage{lipsum}  
\usepackage{multirow}
\usepackage{units}
\usepackage{textcomp} 
\usepackage{smartdiagram}
\usepackage{pgfplots}
\usepackage{balance}
\usepackage{url}
\usepackage[skip=0pt,font=footnotesize]{caption}
\usepackage[pdftex]{hyperref}
\usepackage{adjustbox}
\usepackage{booktabs}
\usepackage{cite}
\usepackage{float}
\usepackage[normalem]{ulem}
\usepackage{cancel}
\usepackage{tabularx}
\usepackage{caption}
\usepackage{longtable}
\usepackage{multicol}
\usepackage{colortbl}
\usepackage{enumitem}
\usepackage{pifont}
\usepackage{float}

\pgfplotsset{compat=1.18}
\makeatletter
\def\IfClass#1#2#3{\@ifundefined{opt@#1.cls}{#3}{#2}}
\makeatother

\IfClass{IEEEtran}{

\newcommand{\citet}[1]{\cite{#1}}
}{}

\definecolor{Ocean}{RGB}{129,194,234}
\definecolor{MotivationColor}{RGB}{211, 222, 190}
\definecolor{Setupcolor}{RGB}{160, 201, 242}
\definecolor{Techniquecolor}{RGB}{255, 200, 208} 
\definecolor{ImportantColor}{RGB}{255, 200, 208} 
\definecolor{Attributescolor}{HTML}{E3D7FF}
\definecolor{lavender}{RGB}{245, 230, 200}
\newcommand{\cmark}{\ding{51}}%
\newcommand{\xmark}{\ding{55}}%

\definecolor{limingcolor}{RGB}{255,165,0}   

\newcommand{\SE}{\mathrm{SE}}
\newcommand{\SO}{\mathrm{SO}}
\newcommand{\etal}{\textit{et al.}}

\begin{document}
\title{Robotic Manipulation via Imitation Learning: Taxonomy, Evolution, Benchmark, and Challenges}
\IfClass{IEEEtran}{
	\author{Zezeng Li$^1$, Alexandre Chapin$^1$, Enda Xiang$^2$, Rui Yang$^1$, Bruno Machado$^1$,  \\ Na Lei$^3$, Emmanuel Dellandrea$^1$, Di Huang$^2$, and Liming Chen$^{1,4}$
	\thanks{1 Ecole Centrale de Lyon, CNRS, Universite Claude Bernard Lyon 1, INSA Lyon, Université Lumière Lyon
2, LIRIS, UMR5205, 69130 Ecully, France}
        \thanks{2 Beihang University, Beijing, China}
        \thanks{3 Dalian University of Technology, Dalian, China}
	\thanks{4 Institut Universitaire de France (IUF)}
    }
	\markboth{IEEE Transactions on Pattern Analysis and Machine Intelligence. Preprint version.}{Li \MakeLowercase{\textit{et al.}}: Robotic Manipulation via Imitation Learning: Taxonomy, Evolution, Benchmark, and Challenges}
}{
	\titlesize{19}
	\author{Zezeng Li$^\dag$}
	\authorshort{Zezeng Li}
	\affiliations{$^\dag$Ecole Centrale de Lyon}
}

\maketitle

\vspace{-3mm}
\begin{abstract}
Robotic manipulation~(RM) is central to enabling autonomous robots to interact with and alter their environments in real-world scenarios. Among the learning paradigms, imitation learning has emerged as a powerful approach, allowing robots to rapidly acquire complex manipulation skills from human demonstrations. This survey provides the first systematic review dedicated to imitation learning for robotic manipulation. We identify and analyze a large set of representative studies selected for their scientific quality and community impact. For each, we provide a structured summary covering purpose, technical implementation, taxonomy, input formats, priors, strengths, limitations, and citation metrics. Beyond cataloging, we trace the chronological evolution of imitation learning techniques within robotic manipulation policies~(RMPs), highlighting key methodological shifts—from diffusion and flow matching to autoregressive and affordance-driven policies. Where available, we compile benchmark results and conduct quantitative comparisons, enabling an integrated view of performance across tasks and environments. Finally, we outline open challenges such as generalization, embodiment diversity, data efficiency, and benchmark standardization, and we discuss promising directions toward scalable and general-purpose RMPs. By synthesizing methods, benchmarks, and challenges, this survey aims to serve both as an entry point for newcomers and a reference for active researchers seeking to advance imitation learning in RMPs.
\end{abstract}


\vspace{-1mm}
\section{Introduction}\label{sec:introduction}
\subsection{Motivation}
\vspace{-1mm}
\textbf{Robotic manipulation (RM)} refers to the ability of robots to physically interact with and transform their surroundings by grasping, moving, assembling, or otherwise altering objects. It is a core capability for deploying autonomous systems in the real world. The importance of manipulation has long been recognized: already Aristotle described the human hand as 
``the tool of tools''~\cite{aristotle_parts}, while Anaxagoras argued that ``man is the most intelligent of the animals 
because he has hands''~\cite{diels_fragments}. \textbf{RM} is crucial because it enables automation of tasks that are too dangerous, precise, repetitive, or labor-intensive for humans, spanning domains such as manufacturing, healthcare, logistics, and household assistance. By extending human reach into hazardous or delicate settings, \textbf{RM} enhances safety, efficiency, and productivity. Achieving robust RM, however, requires the development of robust control policies that allow robots to act adaptively in dynamic and unstructured environments—an enduring challenge at the core of modern robotics.

A \textbf{robotic manipulation policy~(RMP)} specifies how a robot selects and executes actions based on its sensory observations to achieve a manipulation goal, i.e., $\pi:\mathcal{O}\times\mathcal{G}\to\mathcal{A}, a_t=\pi(o_t,g)$. Here 
$\mathcal{O}$ is the observation space, 
$\mathcal{G}$ is the goal space~(e.g., language instructions, object states), and 
$\mathcal{A}$ is the action space.
However, the space of possible object states, contact dynamics, and task variations is combinatorially large, making it impossible to encode effective policies through explicit rules or heuristics. Consequently, current state-of-the-art approaches are predominantly \textbf{data-driven}, leveraging deep learning to learn representations and control policies from large-scale datasets. Yet, even these approaches face significant challenges, since robots must operate in \textbf{dynamic and unstructured environments}, where not only object properties and task goals but also \textbf{environmental conditions}---such as lighting, clutter, occlusions, or background changes---can vary unpredictably. To address these difficulties, \textbf{imitation learning~(IL)} has emerged as a powerful paradigm that learns a policy from expert demonstrations 
$\mathcal{D}=\{ \{(o_t^i, a_t^i)\}_{t=1}^{T_i} \}_{i=1}^N$, with the objective $\pi_\theta \approx \pi_E$, where $\pi_E$ is the expert policy.
Recent progress amplified by advances in computer vision and large language models~(LLMs) has further enhanced robots' ability to perceive, reason, and plan actions. This survey focuses on IL-based RMPs, providing a comprehensive analysis of methodologies, benchmarks, applications, and outlining future directions toward scalable and general-purpose RMPs.


\begin{figure*}[htp!]
\centering
\includegraphics[width=0.9\textwidth]{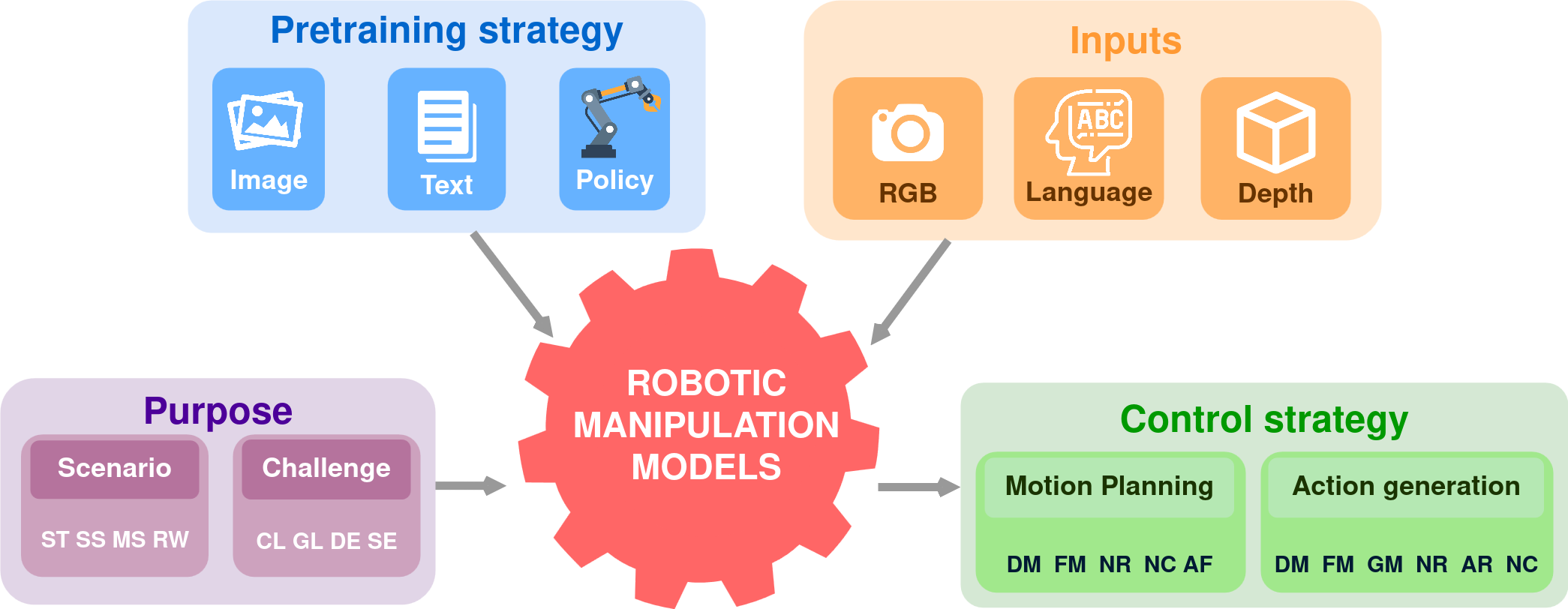}
\caption{Robotic manipulation classification over four perspectives: their purpose, pretraining strategy, input types, and finally their control strategy (DM, FM, GM, NR, AR, NC,
and AF denote diffusion model, flow matching, Gaussian mixture, naive regression, autoregressive, naive classification, and affordance)}
\label{fig:allTaxonomy}
\vspace{-3mm}
\end{figure*}

In light of the fast-paced developments in RMP, this review primarily considers research conducted between 2021 and 2025. Our motivation is \textbf{to provide newcomers to this field with a rapid and comprehensive understanding of the current research landscape, while also supporting active researchers by helping them efficiently locate related works of interest}. By presenting the hierarchical taxonomy, purpose, input, and pre-training strategies, strengths, limitations, and citations of each paper, we aim to distill the core information and enable researchers to quickly assess which approaches align with their specific needs. Furthermore, we present quantitative comparisons across different benchmarks and \textbf{summarize highly sought-after application areas and key existing challenges in the field, intending to mobilize collective efforts within the community to accelerate progress}.

\subsection{Related Surveys}
In recent years, several survey papers on embodied intelligence and robotic manipulation have been published. In this section, we discuss the most relevant ones~\cite{kroemer2021review,liu2021deep,suomalainen2022survey,han2023survey,newbury2023deep,zheng2024survey,sun2025review,sapkota2025vision}. 

In 2021, Kroemer et al.~\cite{kroemer2021review} surveyed a representative subset of publications prior to that year that applied machine learning to RM. They proposed a unified framework by formulating and categorizing existing RM methods based on representation learning, policy learning, and skill transfer. 
Liu \etal~\cite{liu2021deep} and Dong \etal~\cite{han2023survey} reviewed advances in deep reinforcement learning for robotic manipulation control. Markku \etal~\cite{suomalainen2022survey} focus on in-contact tasks that robots can handle and how these tasks are controlled and represented. Newbury \etal~\cite{newbury2023deep} reviewed research on six-degree-of-freedom (6-DOF) grasp synthesis. Sapkota \etal~\cite{sapkota2025vision} surveyed over 80 Vision-Language-Action (VLA) models developed in the past 3 years, highlighting key advancements in this area. Ma \etal~\cite{ma2024survey}, Zhong \etal~\cite{zhong2025survey}, and Din \etal~\cite{din2025vision} also focus on VLA models.

\begin{figure*}[htp!]
\centering
\includegraphics[width=0.75\textwidth]{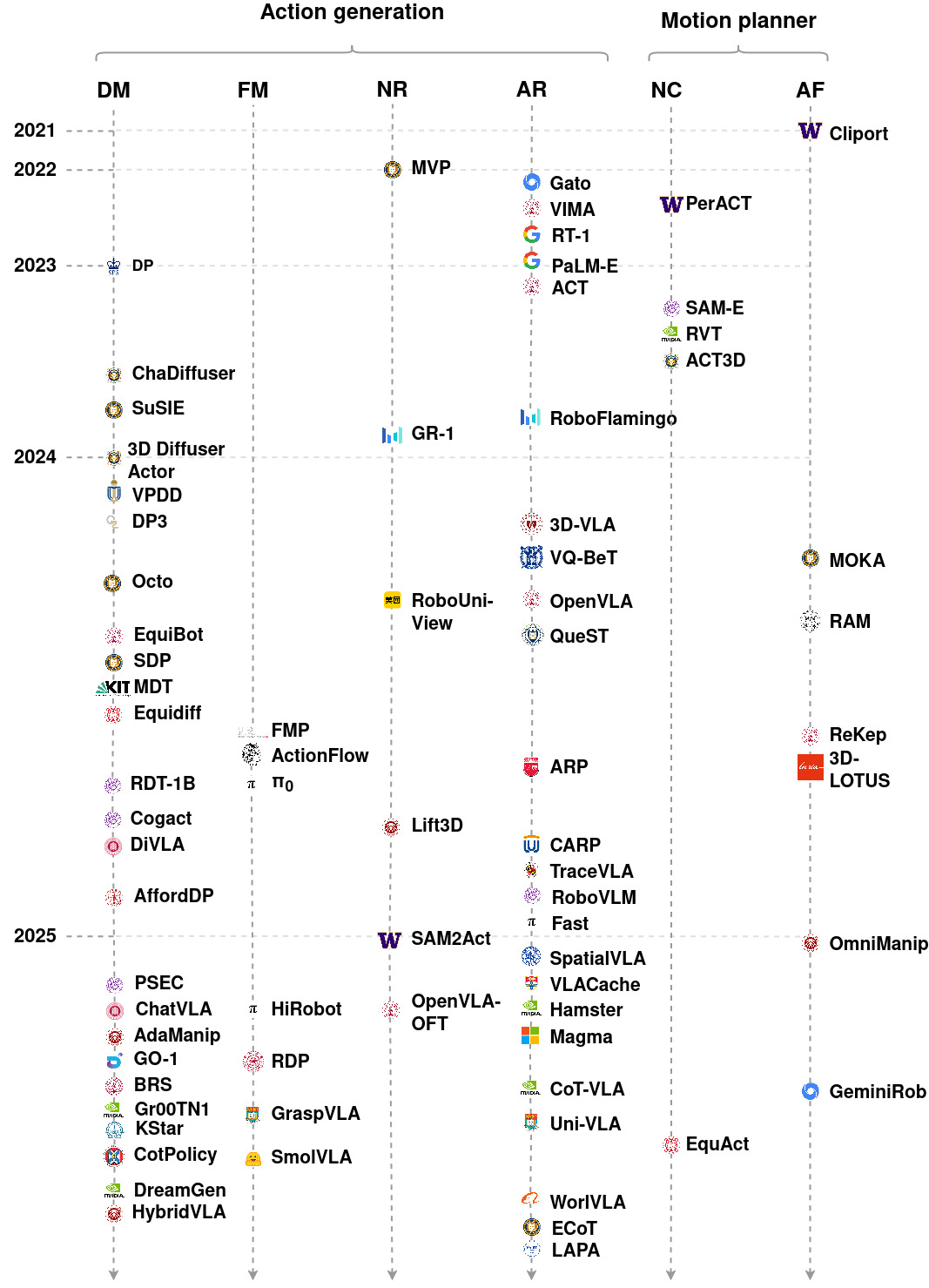}
\caption{Timeline of the models explored in this survey. Each model is classified by its control strategy as presented in Tab.~\ref{tab:taxonomy}.}
\label{fig:timeline}
\vspace{-3mm}
\end{figure*}

Zheng et al.~\cite{zheng2024survey} categorized related work into three main areas: embodied perceptual learning, embodied policy learning, and embodied task-oriented learning. Their survey is the most closely related to ours. However, our work significantly differs in both methodology and information presentation. 1) We \textbf{identified 82 of the most representative papers, selected based on community attention and intrinsic quality}. 2) For each, we \textbf{provide a structured summary covering purpose, technical implementation and hierarchical classification, input formats, key priors, strengths and limitations, and citation metrics}. This format allows researchers to grasp essential information and locate relevant studies quickly.  3) We \textbf{compiled in Fig.~\ref{fig:timeline} the chronological development of key technologies, offering a timeline-based view for understanding the evolution of the field}. 4) Where available, we also \textbf{report benchmark results and provide quantitative evaluations and comparative analyses of existing methods}.

\subsection{Paper Selection Criteria and Content Extraction}

We began by collecting an initial pool of approximately 150 papers through continuous monitoring of the literature and targeted keyword searches. From this pool, we selected a subset of works that were judged to be classic, distinctive, or particularly influential. To quantify influence, we considered both the average monthly citation count and indicators of community attention such as social media engagement. The selected papers were then organized chronologically within each methodological category, as shown in Tab.~\ref{tab:taxonomy} and Tab.~\ref{tab:procons}. Owing to space constraints, we typically retained only one representative paper from each series; for instance, within the RT family of models, we report only RT-1~\cite{brohan2022rt}.

In Tab.~\ref{tab:taxonomy}, we summarized the technique type, target scenarios, challenges, pre-training data, and input data types for each method. The following sections largely expand on the content recorded in Tab.~\ref{tab:taxonomy}. To help readers quickly locate the information of interest, we have organized works of the same category into the same section, separated by horizontal lines. Articles within the same technical area are sorted chronologically. Additionally, the ``Purpose" section highlights the motivation behind each paper, suggesting potential desirable properties of these methods. Tab.~\ref{tab:procons} outlines the core tools used by each method, their pros and cons, and average monthly citations, providing a macro-level evaluation of these approaches.

The remaining sections of this paper are organized according to Fig.~\ref{fig:allTaxonomy} and Tab.~\ref{tab:taxonomy}. First, we give a hierarchical taxonomy of RMP from the perspective of control strategy and introduce the technological evolution for each technique (Section~\ref{sec:strategy}).
Then, we summarize the target challenge and task scenario of each paper (Section~\ref{sec:purpose}), summarize pretraining strategies (Section~\ref{sec:Pretraining}), present relevant benchmarks, evaluation metrics, and conduct performance comparisons and analyses (Section~\ref{sec:evaluation}). Finally, we discuss future challenges and research directions, and conclude the survey (Section~\ref{sec:conclusion}).

\section{Classification based on Control Strategy}\label{sec:strategy}

In this section, we present a hierarchical taxonomy of RMP from the perspective of control strategy~(see Tab.~\ref{tab:taxonomy} for details). We begin by categorizing existing RMP into two main groups: \textbf{action generation} and \textbf{task planner}. Task planners focus on predicting high-level information, such as key poses and affordance maps, 
and enabling manipulation with the aid of motion planning algorithms. In this taxonomy, we highlight how leading RMP approaches effectively integrate generative models, such as diffusion models~\cite{ho2020denoising,song2020denoising,song2020score,li2023dpm}, flow matching~\cite{lipman2023flow}, and autoregressive models~\cite{lutkepohl2013vector,oliva2018transformation}, alongside 

\begin{onecolumn}
{   
    \scriptsize
    \linespread{0.96} \selectfont
    \setlength{\tabcolsep}{3pt}
\begin{longtable}{@{}c|ccccc|cccc|cc|ccccc|cc|c|cc|cccc|cccc@{}}
     \caption{Taxonomy of RMPs. DH, PN, OS, and TX represent depth, proprioception, object state, and text. DM, FM, GM, NR, AR, NC, AF denote diffusion model, flow matching, Gaussian mixture, naive regression, autoregressive, naive classification, affordance.  ID, OD, ST, SS, MS, RW, CL, GL, DE, SE represent in-domain, out-of-domain, single-task, single-scenario, multi-scenario, real-world, continual learning, generalization, data efficiency, sampling efficiency.}\\
    \bottomrule 
    \multicolumn{1}{c}{\multirow{3}[5]{*}{\textbf{Article}}} &\multicolumn{5}{c}{\textbf{Input}}
    &\multicolumn{11}{c}{\textbf{Control strategy}} 
    &\multicolumn{5}{c}{\textbf{Pretraining}} 
    &\multicolumn{8}{c}{\textbf{Purpose}}\\
    \cmidrule(r){2-6} \cmidrule(r){7-17} \cmidrule(r){18-22}  \cmidrule(r){23-30}
    \multicolumn{1}{c}{} &\multicolumn{5}{c}{} &\multicolumn{6}{c}{action generation}&\multicolumn{5}{c}{task planner}  & \multicolumn{2}{c}{images}& \multicolumn{1}{c}{}& \multicolumn{2}{c}{actions} &\multicolumn{4}{c}{scenarios} &\multicolumn{4}{c}{challenges}   \\
    \cmidrule(r){7-12} \cmidrule(r){13-17} \cmidrule(r){18-19} \cmidrule(r){21-22} \cmidrule(r){23-26} \cmidrule(r){27-30}
    \multicolumn{1}{c}{} & 
    \multicolumn{1}{c}{\cellcolor{MotivationColor}RGB} & \multicolumn{1}{c}{\cellcolor{MotivationColor}DH} &  \multicolumn{1}{c}{\cellcolor{MotivationColor}PN}  &  \multicolumn{1}{c}{\cellcolor{MotivationColor}OS} & \multicolumn{1}{c}{\cellcolor{MotivationColor}TX} & \multicolumn{1}{c}{\cellcolor{Techniquecolor}DM} &
    \multicolumn{1}{c}{\cellcolor{Techniquecolor}FM} &
    \multicolumn{1}{c}{\cellcolor{Techniquecolor}GM} &
    \multicolumn{1}{c}{\cellcolor{Techniquecolor}NR} & \multicolumn{1}{c}{\cellcolor{Techniquecolor}AR} & \multicolumn{1}{c}{\cellcolor{Techniquecolor}NC} &\multicolumn{1}{c}{\cellcolor{Setupcolor}DM}&  \multicolumn{1}{c}{\cellcolor{Setupcolor}FM} &
    \multicolumn{1}{c}{\cellcolor{Setupcolor}NR} &
    \multicolumn{1}{c}{\cellcolor{Setupcolor}NC} &\multicolumn{1}{c}{\cellcolor{Setupcolor}AF}  & \multicolumn{1}{c}{\cellcolor{lavender}ID}& \multicolumn{1}{c}{\cellcolor{lavender}OD} & \multicolumn{1}{c}{\cellcolor{lavender}TX} & \multicolumn{1}{c}{\cellcolor{lavender}ID}& \multicolumn{1}{c}{\cellcolor{lavender}OD} &\multicolumn{1}{c}{\cellcolor{Attributescolor}ST} 
    &\multicolumn{1}{c}{\cellcolor{Attributescolor}SS} 
    &\multicolumn{1}{c}{\cellcolor{Attributescolor}MS} 
    &\multicolumn{1}{c}{\cellcolor{Attributescolor}RW}  &\multicolumn{1}{c}
    {\cellcolor{Attributescolor}CL}&\multicolumn{1}{c}
    {\cellcolor{Attributescolor}GL}&\multicolumn{1}{c}
    {\cellcolor{Attributescolor}DE} 
    &\multicolumn{1}{c}
    {\cellcolor{Attributescolor}SE}  \\
    \hline 
    \endhead\label{tab:taxonomy}
    DP~\cite{chi2023diffusion}& \cellcolor{MotivationColor}\checkmark &\cellcolor{MotivationColor}& \cellcolor{MotivationColor}\checkmark& \cellcolor{MotivationColor} &\cellcolor{MotivationColor} &\cellcolor{Techniquecolor}\checkmark &\cellcolor{Techniquecolor}  &\cellcolor{Techniquecolor}&  \cellcolor{Techniquecolor} &\cellcolor{Techniquecolor}  &\cellcolor{Techniquecolor} &\cellcolor{Setupcolor} &\cellcolor{Setupcolor} &      
    \cellcolor{Setupcolor} &\cellcolor{Setupcolor} &       
    \cellcolor{Setupcolor}& \cellcolor{lavender}& \cellcolor{lavender} & \cellcolor{lavender}& \cellcolor{lavender} &\cellcolor{lavender} &\cellcolor{Attributescolor}&\cellcolor{Attributescolor}\checkmark&\cellcolor{Attributescolor} &\cellcolor{Attributescolor}&\cellcolor{Attributescolor}&\cellcolor{Attributescolor} &\cellcolor{Attributescolor}&\cellcolor{Attributescolor}\\
    SuSIE~\cite{black2023zero}& \cellcolor{MotivationColor}\checkmark &\cellcolor{MotivationColor}& \cellcolor{MotivationColor}& \cellcolor{MotivationColor} &\cellcolor{MotivationColor}\checkmark &\cellcolor{Techniquecolor}\checkmark &\cellcolor{Techniquecolor}  &\cellcolor{Techniquecolor}& \cellcolor{Techniquecolor} &\cellcolor{Techniquecolor}  &\cellcolor{Techniquecolor} &\cellcolor{Setupcolor} &\cellcolor{Setupcolor} &      
    \cellcolor{Setupcolor} &\cellcolor{Setupcolor} &       
    \cellcolor{Setupcolor}& \cellcolor{lavender}& \cellcolor{lavender} & \cellcolor{lavender}& \cellcolor{lavender} &\cellcolor{lavender} &\cellcolor{Attributescolor} &\cellcolor{Attributescolor}\checkmark&\cellcolor{Attributescolor} &\cellcolor{Attributescolor}&\cellcolor{Attributescolor} &\cellcolor{Attributescolor}\checkmark &\cellcolor{Attributescolor}&\cellcolor{Attributescolor}\\
\bottomrule
    \end{longtable}
 }
\end{onecolumn}
\twocolumn

\noindent naive regression and classification with multi-layer perceptron (MLP). We also discuss the evolution of each technique. Further elaborations are provided in the subsections that follow.


\subsection{Policy for Action Generation}\label{sec:Actiongeneration}

For action generation methods, we further classify them based on the type of action produced: \textbf{continuous} or \textbf{discrete}.

\subsubsection{\textbf{Continuous Action}}
For continuous action generation, we categorize the methods into several types: \textbf{diffusion model}-based, \textbf{flow matching}-based, and \textbf{naive regression}-based. 

\paragraph{\textbf{Action generation via diffusion models}} 
The progression of this area has evolved from the initial diffusion policy to 3D diffusion policy, followed by the introduction of equivariant diffusion policy. This was further enhanced with the combination of diffusion and autoregressive policies, leading to the emergence of the test-time diffusion policy.


Diffusion Policy (DP) ~\cite{chi2023diffusion} pioneered this idea by formulating control as a conditional denoising process, iteratively refining actions from noise given current observations. Building on this, SuSIE~\cite{black2023zero} used a pre-trained image-editing diffusion model for visual subgoal generation, paired with a controller to execute these goals, enabling zero-shot generalization. ChaDiffuser~\cite{xian2023chaineddiffuser} and VPDD~\cite{he2025learning} incorporated autoregressive planning and predictive video modeling to extend diffusion into longer-horizon control. Spatial reasoning was enhanced in 3D Diffusion Policy (DP3)~\cite{ze20243d} and 3D Diffuser Actor ~\cite{3d_diffuser_actor} by integrating point clouds and 3D scene features, improving robustness to viewpoint and object variation. To exploit structural priors, EquiDiff~\cite{wang2024equivariant} and EquiBot~\cite{yangequibot} embedded $\SO$(2) or $\SE$(3) equivariance into diffusion, yielding higher data efficiency and invariance to rotations and translations.

Recent work has also targeted continual learning, adaptability, and domain transfer. SDP~\cite{wangsparse} accelerated sampling with reduced denoising steps; PSEC~\cite{liu2025skill} represented skills in parameter space via LoRA-style modules, allowing incremental expansion and composition without retraining the whole model. AffordDP~\cite{wu2024afforddp} guided sampling with transferable 3D affordances to enable cross-category generalization, while KStar~\cite{lv2025spatial} enforced kinematics-aware planning for collision-free dual-arm motions. AdaManip~\cite{wangadamanip} adapted to articulated objects with configuration-dependent control. These advances have moved diffusion policies from lab demonstrations to real-world deployments, such as BRS~\cite{jiang2025behavior} for household tasks and CoTPolicy~\cite{sochopoulos2025fast}, which integrates chain-of-thought planning with diffusion to tackle complex multi-instruction, long-horizon tasks. Collectively, these works expand diffusion policies into versatile and semantically aware controllers.


In parallel, foundation models have combined diffusion with large-scale vision-language learning to enable generalist robot behavior. Octo~\cite{team2024octo} trained on 800k diverse demonstrations, supporting multi-modal commands and multi-embodiment control via transformer-based diffusion decoding. DiVLA~\cite{wen2024diffusion} unified vision-language model (VLM) reasoning with a diffusion control head, translating high-level instructions into low-level actions. CogACT~\cite{li2024cogact} decoupled perception/reasoning from a learned diffusion transformer for improved adaptability, while ChatVLA~\cite{zhou2025chatvla} staged training between a diffusion controller and a pre-trained VLM, producing agents that both understand instructions and execute precise actions. Scaling has amplified capabilities. RDT-1B~\cite{liu2024rdt}, with 1.2B parameters and over a million demonstrations, achieved state-of-the-art performance on complex bimanual skills. GO-1 ~\cite{bu2025agibot} demonstrated cross-embodiment generalization across hundreds of robots and tasks by learning a unified latent action planner. GR00T-N1~\cite{bjorck2025gr00t} paired a frozen VLM for semantic reasoning with a learned diffusion executor for reactive control, improving interpretability and task decomposition. Data augmentation has also been explored: DreamGen~\cite{jang2025dreamgen} synthesized novel demonstrations via generative world and action models to enhance robustness. Finally, HybridVLA~\cite{liu2025hybridvla} combined autoregressive reasoning for long-horizon planning with diffusion for precise action generation, outperforming single-paradigm baselines. 

\paragraph{\textbf{Action generation via flow-matching}}
Flow-matching, as a novel generative model, has garnered significant attention due to its ability to generate high-quality results more quickly compared to diffusion models. Many researchers have explored its potential in combination with robotic action generation, leading to the emergence of several outstanding works in this area. Initially, FMP~\cite{zhang2024affordance} gave the first attempt to ground vision-language model (VLM) affordance with flow matching for robot manipulation, enabling more stable training and faster inference. Reactive diffusion policy (RDP)~\cite{xue2025reactive}, a
novel slow-fast visual-tactile imitation learning algorithm, allowing robots to adjust their policies in real-time for contact-rich manipulation tasks. ActionFlow~\cite{funk2024actionflow} introduces an $\SE$(3) invariant Transformer, which enables informed spatial reasoning based on the relative $\SE$(3) poses between observations and actions. Later, $\pi_0$~\cite{black2410pi0} integrates pre-trained VLM with flow matching to inherit Internet-scale semantic knowledge, enhancing the robot's ability to efficiently learn and execute a wide variety of tasks, such as laundry folding, table cleaning, and assembling boxes. GraspVLA~\cite{deng2025graspvla} integrates autoregressive perception tasks and flow-
matching-based action generation into a unified chain-of-thought process, enabling joint training on synthetic action data and Internet semantics data, exhibiting direct sim-to-real transfer
and strong zero-shot generalization, as well as few-shot adaptability to specialized scenarios and human preferences. This was further advanced by HiRobot~\cite{shi2025hi}, which utilizes VLMs for both high-level reasoning and low-level task execution, enabling the processing of much more complex prompts. Most recently, SmolVLA~\cite{shukor2025smolvla} focused on making these techniques more efficient in resource-constrained environments, incorporating lightweight visual learning systems while maintaining the robustness of flow matching. Overall, these developments illustrate the continuous refinement of flow-matching techniques, expanding their applicability from basic manipulation to more complex, adaptable, and resource-efficient systems.


\paragraph{\textbf{Action generation via naive regression}}

Some researchers, particularly those working with the VLA model, argue that the visual and language encoders are responsible for the core task of information extraction, a function that is universal across tasks. In this view, the action policy head merely needs to predict the actions based on the extracted features, with different scenarios potentially requiring fine-tuning of the action policy head. As a result, they suggest that the action generation module does not require a complex architecture. Instead, they opt for a simple MLP and use straightforward regression techniques to learn action generation.

The evolution of action generation via naive regression has progressed through several key stages. Initially, MVP~\cite{xiao2022masked} introduced a masked vision-policy framework, which regressed action commands directly from visual features. This approach demonstrated that self-supervised visual pre-training can be effective for motor control. GR-1~\cite{wuunleashing} generative video models to predict future frames and behaviors, and shows that such pretraining benefits downstream robot manipulation when fine-tuned. RoboUniView~\cite{liu2024robouniview} further advanced the model by integrating multi-view vision systems, addressing the challenge of being sensitive to camera specifications and positions. Lift3D~\cite{jia2024lift3d} then extended this approach by incorporating 3D depth information, improving accuracy in spatially complex environments. SAM2Act~\cite{fang2025sam2act} equips the agent with spatial memory awareness, allowing it to solve spatial memory-based tasks. Most recently, OpenVLAOFT~\cite{kim2025fine} explored crucial design choices for adapting VLAs, including various action decoding schemes, action representations, and learning objectives for fine-tuning. Using OpenVLA as the base model, this work provides valuable insights on how to fine-tune VLAs for novel setups. Together, these advancements illustrate a shift from simple regression to more adaptable and robust approaches in robotic action generation.


\subsubsection{\textbf{Discrete Action}}
Discrete action generation is a crucial component of RMP, where the goal is to predict and execute a finite set of distinct actions based on sensory inputs. Its advantage lies in reducing the solution space and simplifying the problem-solving process. But it cannot provide fine-grained, continuous manipulation actions. Discrete action generation methods are generally categorized into two main approaches: \textbf{autoregressive}-based methods and \textbf{naive classification}-based methods. Autoregressive models generate actions sequentially, conditioning actions on the previous, while naive classification models treat the problem as a classification task, directly mapping inputs to discrete actions.

One of the key advantages of discrete action generation is its efficiency. Since actions are categorized into distinct classes, the learning process benefits from the vast array of classification techniques available in deep learning, enabling faster and more stable training. These methods can also leverage well-established classification models and frameworks, making them particularly well-suited for tasks with clear, finite action spaces. However, a notable disadvantage is the lack of precision. Since discrete action generation typically relies on predefined action categories, it may struggle with tasks that require fine-grained, nuanced control. The discrete nature of these models limits their ability to handle complex or highly variable tasks that require more detailed action specifications.

This section explores both autoregressive-based and naive classification-based approaches, emphasizing their unique characteristics, challenges, and key advancements in the field.


\paragraph{\textbf{Discrete action generation via autoregressive model}}

Given the success of Transformer-based networks in LLMs and VLMs, an increasing number of studies have integrated the language modality into visuomotor policies, establishing a paradigm of autoregressive action generation conditioned on multimodal inputs.
Gato~\cite{reed2022generalist} firstly frames a single large Transformer as a generalist embodied agent able to produce multiple output modalities from diverse inputs, which demonstrates the feasibility of weight-sharing across tasks and embodiments while highlighting limits in scaling and task-specific performance. 
VIMA~\cite{jiang2022vima} extends the prompting paradigm to robotics by using multimodal prompts to specify tasks; it introduces a desktop-task suite and shows that a Transformer trained with multimodal prompts generalizes well in few-shot settings. 
RT-1~\cite{brohan2022rt} presents an engineering recipe for tokenizing all inputs outputs and training a large end-to-end Transformer to generate actions directly from raw observations.
It's worth noting that although the RT-1 model is small, it establishes the VLA concept; RT-2~\cite{brohan2023rt} later explicitly defines and popularizes it, and this has had a profound impact on subsequent work.
PaLM-E~\cite{driess2023palm} injects visual and state embeddings into a large language model to form a single embodied multimodal model, showing that joint multimodal training yields positive transfer: the model can perform perception, multi-step reasoning, and control-related tasks within the same architecture.
RoboFlamingo~\cite{li2023vision} demonstrates that large vision–language foundation models can be repurposed as effective imitation policies: by adding lightweight policy heads and small amounts of tuning, VLMs can interpret multimodal demonstrations and drive robot behavior with minimal modification.
OpenVLA~\cite{kim2024openvla} provides an open-source VLA model trained on large-scale vision-language-action datasets, with a unified architecture for perception, reasoning, and control. It aims to democratize research in generalist policies and offers tools for fine-tuning and evaluation.
RoboVLM~\cite{li2024towards} provides a systematic empirical study on the choices that most affect VLA model performance, covering dataset composition, architecture design, and training strategies. It offers guidelines for building more capable generalist robot policies.

Using autoregressive models in single-task or single-scenario settings is another widely studied approach.
ACT~\cite{zhaolearning2023} package high-frequency, low-level controls into discrete “action chunks,” greatly shortening autoregressive sequence lengths and improving sample efficiency in low-demo regimes.
VQ-BeT~\cite{lee2024behavior} proposes representing actions in a compact latent space learned from demonstrations, which allows for smooth, temporally consistent behavior generation. This reduces the complexity of action prediction and improves generalization across different manipulation tasks. 
QueST~\cite{mete2025quest} introduces a method for learning skill abstractions in a self-supervised manner, where low-level control policies are organized into reusable high-level skills. 
ARP~\cite{zhang2025autoregressive} introduces variable-step or mixed-token prediction mechanisms that let a single architecture emit tokens at different frequencies and granularities, reducing the number of autoregressive steps while increasing adaptability.
CARP~\cite{gong2024carp} extends autoregressive policy learning with a coarse-to-fine prediction scheme, where a high-level action plan is refined into fine-grained motor commands. This structure improves efficiency and accuracy in manipulation.

Subsequent works often improve the quality of action generation from different directions; one approach is to enhance the model’s spatial perception capabilities.
3D-VLA~\cite{zhen20243d} introduces a generative world model that jointly learns 3D visual perception, language understanding, and action generation by integrating 3D scene representations into a vision-language-action architecture.
Based on OpenVLA, TraceVLA~\cite{zheng2024tracevla} introduces visual trace prompting, which augments inputs with visual trajectory cues from past frames to improve spatial-temporal grounding. This boosts generalization and performance of VLA models on multi-step manipulation tasks.
SpatialVLA~\cite{qu2025spatialvla} introduces spatially aware representations include Ego3D position encoding and adaptive action grids, to inject 3D robot-agnostic spatial structure into VLA models, showing improved zero-shot and transfer performance after pretraining on millions of real robot episodes.
Hamster~\cite{li2025hamster} argues for a hierarchical VLA design where a high-level VLM predicts coarse 2D trajectories and a low-level 3D controller executes motions. This separation improves domain transfer and yields large gains over monolithic VLA baselines.

Meanwhile, because robot demonstration data are scarce, several works have begun exploring how to learn useful action representations from large-scale actionless videos.
LAPA~\cite{ye2024latent} proposes unsupervised latent-action pretraining from actionless videos by learning discrete latent actions and then finetuning to map latents to robot actions, enabling large-scale pretraining without costly robot action labels.
UniVLA~\cite{bu2025univla} learns task-centric latent action representations from heterogeneous, cross-embodiment videos so a single policy can be decoded onto different robots, enabling scalable, compute-efficient pretraining that transfers across embodiments and environments.
WorldVLA~\cite{cen2025worldvla} unify a discrete autoregressive world model and an action model in one VLA framework, so images and actions are jointly modeled. The coupled model improves video prediction and action performance, and introduces masking attention strategies to reduce autoregressive action error accumulation.

Some works aim to improve model encoding and inference efficiency, since the generation frequency is also critical for manipulation tasks.
Fast~\cite{pertsch2025fast} proposes a compression-based action tokenization that transforms short action chunks into frequency space using the discrete cosine transform, then compresses those coefficients with byte-pair encoding, producing compact tokens that better represent high-frequency.
VLACache~\cite{xu2025vla} proposes an adaptive token-caching mechanism that detects unchanged visual tokens across timesteps and reuses their key-value computations, reducing VLA inference cost with minimal loss in success rate.

Other works also introduce the concept of chain-of-thought~(CoT) reasoning from LLMs, achieving promising results as well.
ECoT~\cite{zawalskirobotic} trains VLAs to perform embodied, multi-step reasoning like plans, sub-tasks and visually grounded waypoints before predicting actions, showing large improvements in robustness and interpretability without extra robot data.
CoT-VLA~\cite{zhao2025cot} integrates explicit visual CoT by autoregressively predicting future image frames as intermediate visual goals before generating short action sequences. This approach improves temporal planning and yields substantial gains on real and simulated manipulation benchmarks.

\paragraph{\textbf{Discrete action generation via naive classification}}
The evolution of discrete action generation via naive classification can be observed across the following works: HULC~\cite{mees2022hulc} was one of the first to apply naive classification for discrete action generation. It simplifies the action generation process by treating it as a classification task, enhancing the efficiency for robotic manipulation. BridgeVLA~\cite{li2025bridgevla} expanded on this approach by integrating VLAs. It unifies visual and language representations to better contextualize actions within dynamic environments. To leverage the structural priors of the 3D input, BridgeVLA is trained to predict 2D heatmaps, which facilitate more accurate translational action prediction.
VPDD~\cite{he2025learning} further advances the field by introducing a discrete diffusion model to combine generative pre-training on human videos and discrete diffusion policy fine-tuning on a small number of action-labeled robot videos. VPDD emphasizes learning to predict not only actions but also the conditions under which specific actions should be executed, improving the model's generalization ability to new tasks.

\subsection{Policy for Task Planner}\label{sec:planning}
For methods leveraging task planners, we further classify them based on method's type used to produce the targets: generative models, naive classification or affordance prediction. 

\paragraph{\textbf{Task planner via generative models}}
There are two primary generative models which has been integrated into task planner-based robotic manipulation, i.e., diffusion models and flow-matching models. 


SEDF~\cite{urain2023se} was one of the pioneering works that combined diffusion models with robotic manipulation. It decouples the grasp pose selection and the motion planning, resolves the grasp and motion planning problem by iteratively improving the trajectory to jointly minimize the object-grasp diffusion cost and the task-related costs. A0~\cite{xu2025a0} further advanced this approach by introducing a hierarchical, affordance-aware diffusion model. This model decomposes the manipulation task into high-level spatial affordance understanding and low-level action execution, significantly enhancing generalization capabilities and offering an embodiment-agnostic design, making it adaptable to a wide range of robotic systems.

Building on flow matching, FlowMS~\cite{rouxel2024flow} introduces an imitation learning architecture designed for multi-support manipulation tasks, enabling a multi-contact whole-body controller. Leveraging flow matching, it enhances the efficiency of generating feasible manipulation trajectories.

\paragraph{\textbf{Task planner via naive classification}}


PerAct~\cite{shridhar2023peract} is one of the earliest motion-planning-based robotic manipulation policies. It takes a language goal and a voxel grid reconstructed from RGB-D sensors as input, and uses MLPs as the policy head to predict discrete actions, which are then executed using a motion planner. Given the computational burden introduced by voxel-based 3D representations, RVT~\cite{goyal2023rvt} optimized PerAct's network architecture by proposing a multi-view Transformer, reducing complexity while maintaining performance. Act3D~\cite{gervet2023act3d} extends this by lifting 2D pre-trained features to 3D using sensed depth, and predicts the 3D location of the end effector through classification of 3D points in the robot's workspace. SAM-E~\cite{zhang2024sam} incorporated SAM~\cite{kirillov2023segment} as the foundation model for extracting task-relevant features, then framed action prediction as a classification problem, guided by heatmaps, to improve generalization in few-shot adaptations to new tasks. Most recently, EquAct\cite{zhu2025equact} leveraged $\SE$(3) equivariance as a key structural property shared by both policy and language, enhancing generalization to novel 3D scene configurations. This progression reflects a shift from simpler action representations to increasingly sophisticated frameworks that are more generalizable and efficient.

\paragraph{\textbf{Task planner via affordance prediction}}


In robotic manipulation, affordance refers to the potential actions or interactions that an object allows a robot to perform. It represents the relationship between an object and the robot, essentially describing how a robot interacts with objects in its environment. Affordance can be expressed in various forms, such as keypoint~\cite{kuang2024ram,liu2024moka,huangrekep}, segmentation masks~\cite{garcia2024towards,zhang2024sam,fang2025sam2act}, and affordance maps~\cite{shridhar2022cliport,wu2024afforddp,team2025gemini}. By recognizing and utilizing affordances, robots can make more informed decisions about how to manipulate objects, even in dynamic or unstructured environments. This approach reduces the need for exhaustive task-specific programming, allowing robots to generalize across different scenarios, handle a wider range of objects, and perform more complex manipulation tasks.

The evolution of manipulation via motion planning with affordance techniques progresses from Cliport~\cite{shridhar2022cliport}, which formulates tabletop rearrangement as a series of language-conditioned affordance predictions, and thus benefits from the strengths of data-driven scale and generalization.
RAM~\cite{kuang2024ram} refined this approach by introducing a retrieve-and-transfer framework for zero-shot robotic manipulation. It hierarchically retrieves the most similar demonstration from the affordance memory and transfers such out-of-domain 2D affordance to in-domain 3D executable affordance in a zero-shot and embodiment-agnostic manner. MOKA~\cite{liu2024moka} advanced affordance-driven manipulation by incorporating a compact point-based representation of affordance, bridging the VLM's predictions on observed images and the robot's actions
in the physical world. 
By representing a manipulation task as a sequence of Relational Keypoint Constraints, ReKep~\cite{huangrekep} introduced a hierarchical optimization
procedure to solve for robot actions.
3D-LOTUS~\cite{garcia2024towards} focused on improving generalization by integrating foundation VLMs for task planning. It decomposes tasks into step-by-step plans and grounds objects by localizing them with segmentation masks. Once the objects are grounded and the primitive actions are identified, 3D-LOTUS serves as a motion controller to generate action trajectories. GeminiRob~\cite{team2025gemini} utilizes Gemini 2.0 for open-world affordance prediction, enhancing its ability to generalize to unseen environments. This progression reflects a shift toward a more adaptive and generalist agent capable of handling dynamic, complex real-world environments.

\section{Classification based on Purpose}\label{sec:purpose}
This section outlines the motivation behind each paper, highlighting potential desirable properties of the methods. We have primarily classified the works from two views: scenarios and challenges. We aim to help readers quickly find relevant works based on specific scenarios or challenges.

\subsection{Policy for Single Task}\label{sec:singletask}
Early advances in robotic manipulation policy often targeted single-task settings, where the objective was to master a specific manipulation skill efficiently and reliably. 
CLIPort~\cite{xue2025reactive} uses CLIP-extracted semantic visual features to condition pixel-level pick-and-place/transport policies, combined with transport/affordance prediction for sample-efficient single-task learning.
ACT~\cite{xue2025reactive} packs high-frequency, low-level controls into action chunking and performs autoregression at the chunk level, substantially shortening sequence length and improving stability and success rates for fine-grained single tasks with few demonstrations.
Diffusion Policy~\cite{xue2025reactive} uses conditional diffusion models to generate continuous actions or trajectories, excelling at modeling multi-modal target distributions and producing high-quality, natural action sequences. In single-task settings it often yields robust and diverse solutions.
Equivariant Diffusion Policy~\cite{wang2024equivariant} Incorporates SO(2) equivariance into diffusion policies, leveraging task symmetries to improve sample efficiency, generalization, and robustness to observation transforms for single task.
Reactive Diffusion Policy~\cite{xue2025reactive} places diffusion generation inside a closed-loop control loop, repeatedly resampling and adapting actions based on the latest tactile/force inputs. It performed impressively on a single contact-rich task. Collectively, these studies systematically explore optimal architectures, training regimes, and inference pipelines for single-task control, thereby providing a robust foundation for the development of generalist policies that can scale to multiple tasks and diverse environments.

\subsection{Policy for Multi-Task Under Single Scenario}\label{sec:singlescenario}
While model advanced, new iterations created models trained on several tasks but limited to a specific scenario. Some works leverage stronger spatial perception and flexible conditional generation to cover different tasks in same environment. MDT~\cite{reuss2024multimodal} and 3D Diffusion policy~\cite{ze20243d}
emphasize high-quality, long-horizon generation conditioned on multimodal and 3D inputs. PerAct~\cite{shridhar2023peract} emphasizes extreme sample efficiency for many task variants within one scene. RVT~\cite{goyal2023rvt} uses multi-view aggregation and view synthesis within a Transformer architecture to build scalable, high-fidelity 3D perception, achieving fast training and inference while maintaining precision. Other works centers on optimization-driven approaches, using visual relational keypoint constraints to structure tasks so that action sequences can be produced by optimization—yielding interpretable behavior and strong zero-shot adaptability. VoxPoser~\cite{huang2023voxposer} grounds LLM reasoning and constraint into spatial value functions and then uses planning to produce trajectories instead of training policies to directly output actions. ReKep~\cite{huangrekep} models tasks using relational keypoint constraints, framing task execution as a set of optimizable constraints.
Overall, these works, whether because of limitations in problem formulation and nongeneral experimental settings, or because they are constrained by model capacity and the diversity of training data, have not demonstrated task completion across different tasks and scenes.

\subsection{Policy for Multi Scenario}\label{sec:multiscenario}
A few years back, Gato~\cite{reed2022generalist} initiated the idea of creating a multi-purpose ``generalist agent", capable of solving a set of tasks, even tasks it has never seen. A main issue to solve this problem is access to reliable data. Recently, the scale of robotic data has increased by a large margin thanks to community initiatives such as Open-X~\cite{o2024open}. This permitted new models to emerge by getting inspiration from other fields such as natural language processing and computer vision for pretraining~(see Section~\ref{sec:Pretraining}). Leveraging the reasoning capacities of pretrained Visual Language Models, a set of new methods was introduced to obtain agents capable of understanding and reasoning over scenes before acting, leading to \textbf{multi-scenario} policies capable of generalizing. One of the first iteration of such model is RT-1~\cite{brohan2023rt} which pretrains a visual model with text conditioning on a large-scale dataset of tasks. Following work leverages pretrained models on out-of-domain data for their strong reasoning capabilities to learn an action model, which leads to the set of VLAs.

RT-2~\cite{brohan2023rt} Discretizes actions into tokens and jointly pretrains/fine-tunes on large-scale vision–language (VQA-style) data together with real trajectory data, thereby transferring web-scale semantic and reasoning capabilities into multi-task, cross-scene robot control and achieving strong zero-/few-shot generalization and semantic reasoning performance.
OpenVLA~\cite{kim2024openvla} Provides an open-source VLA pretrained on nearly a million real robot demonstrations, emphasizing fast, parameter-efficient fine-tuning for new robots and tasks; it champions large-data-driven generality, reproducibility, and adaptability in cross-scene, multi-task settings.
HiRobot~\cite{shi2025hi} uses a hierarchical inference–execution pipeline that separates high-level language reasoning/task decomposition from low-level real-time control and supports integration of human feedback and online corrections during execution, improving adaptability and robustness on open-ended, multi-step, interactive tasks.
CoT-VLA~\cite{zhao2025cot} introduces a visual chain-of-thought by autoregressively predicting future image frames as intermediate visual goals before generating short action sequences, enhancing temporal reasoning, stepwise planning, long-horizon consistency, and model interpretability across multi-step and multi-scene tasks.
SwitchVLA~\cite{li2025switchvla} proposes an execution-aware task-switching framework that treats task switches as behavior-modulation problems conditioned on execution state and instruction context; by segmenting demonstrations into contact phases and training multi-behavior conditional policies, it improves smoothness and robustness when switching tasks in dynamic and interactive environments.

\subsection{Policy for Real-world Scenario}\label{sec:realscenario}
Currently, while most works incorporate real-world validation, the following efforts specifically optimize for real-world scenarios, making targeted improvements in data scale, training strategies, and data standardization, and demonstrating remarkable performance on real-world manipulation tasks.
RT-2~\cite{brohan2023rt} treats actions as discrete tokens and jointly trains and fine-tunes on VQA-style visual-language data together with real-world trajectory data, successfully transferring the capabilities of large VLMs to real-world robotic tasks.
$\pi_0$~\cite{black2410pi0} focuses on combining pretrained vision–language knowledge with continuous, high-frequency action generation: it uses flow-matching-based continuous action modeling to preserve high control rates and action smoothness, yielding better consistency and frequency responsiveness in semantically guided real-world manipulation.
HiRobot~\cite{shi2025hi} proposes a hierarchical inference–execution pipeline that separates high-level language reasoning and task planning from low-level real-time control; this design improves adaptability to open-ended, multi-step, and interactive real-world tasks and facilitates handling dynamic scenes and online instruction corrections.
Geminiobot~\cite{team2025gemini}, built on the large multimodal foundation model Gemini, applies powerful language understanding, spatial reasoning, and generative capabilities directly to the robot’s closed-loop perception–planning–execution stack, exploiting large models’ strengths in complex reasoning (e.g., multi-step fine manipulations and tool use).
BRS~\cite{jiang2025behavior} focuses on whole-body household scenarios and develops corresponding hardware and data-collection pipelines suited for long-horizon tasks, dual-arm coordination, and constrained-space manipulation. It emphasizes standardization and systematization of large-scale real demonstrations, demonstrating the robustness and reproducibility of end-to-end policies on household tasks.

\subsection{Policy for Continual Learning}\label{sec:continuallearning}
Across recent robot manipulation work, Continual Learning is pursued via complementary policy designs. LOTUS~\cite{wan2024lotus} discovers reusable visuomotor skills from raw demos and composes them with a meta-controller, yielding a continually expanding skill library for lifelong imitation. SDP~\cite{wangsparse} embeds sparsity into diffusion policies via MoE, activating only a few experts per task to curb interference, retain past skills as tasks accrue. PSEC~\cite{liu2025skill} treats skills as plug-and-play LoRA modules in parameter space, enabling iterative expansion and direct skill composition with lightweight routing. ChatVLA~\cite{zhou2025chatvla} tackles forgetting and task interference when unifying perception–language understanding with control, using phased alignment and MoE to co-train a generalist, extensible VLA policy. All four decouple what to reuse (skills/experts/adapters) from how to compose (router/meta-controller), updating only lightweight modules instead of rewriting the whole controller to reduce cross-task interference.

\subsection{Policy for Generalization}\label{sec:generalization}
Currently, robot manipulation learning based on imitation learning suffers from evident overfitting, and the generalization of the policies remains a widespread and unresolved issue. Generalization itself is a broad problem, and researchers have attempted to enhance it from various perspectives, which we categorize into the following two main areas.

\begin{itemize}[leftmargin=20pt]
\item \textbf{Intra-task generalization} demands that a learned policy remain robust under a wide range of contextual shifts, such as changes in object appearance, shape, pose, lighting, distractors, background, or embodiment. 

\item \textbf{Inter-task generalization} requires robots not just to recall previously learned behaviors, but to systematically accumulate, organize, and recombine those skills to perform entirely new tasks.   
\end{itemize}

For example, intra-task generalization implies that a model trained for the task “pick up the mug and place it on the shelf” should be able to handle different instances of mugs (tall vs. short, ceramic vs. plastic, with or without handles), varying poses (mug upright, on its side, or partially occluded), different environments (e.g., office, home), and potential distractors (e.g., cups, bowls nearby). In contrast, inter-task generalization means a model trained on tasks such as: a) picking up the mug and placing it on the shelf, and b) pushing the red block to the left pad, should be able to generalize to a novel task: d) stacking the blue bowl on top of the green plate. In this case, the verb "stack" was not seen during training, the new object pair (bowl, plate) introduces unique object interactions and stability constraints, and the goal condition (achieving vertical alignment for stacking) differs both semantically and physically from the previous tasks.

For \textbf{intra-task generalization}, Cliport~\cite{shridhar2022cliport}, GR-1~\cite{wuunleashing}, and RoboFlamingo~\cite{li2023vision} can generalize across instances with varying colors and positions. VIMA~\cite{jiang2022vima} and PaLM-E~\cite{driess2023palm} demonstrate ability to generalize to new combinations of objects. MVP~\cite{xiao2022masked} disentangles shape and color, enabling it to handle different object geometries and colors. SuSIE~\cite{black2023zero} introduces a goal-conditioned policy based on future observation prediction. All works tackle intra-task generalization from the perspective of objects, while others focus on addressing generalization from the perspective of the scene. RAM~\cite{kuang2024ram} generalizes across various objects and environments using a retrieval-based transfer paradigm. 3D-LOTUS~\cite{garcia2024towards} and Lift3D~\cite{jia2024lift3d} enhance generalization across different instances, background scenes, and lighting conditions. RoboUniView~\cite{liu2024robouniview} maintains high performance under unseen camera parameters by utilizing a unified view representation. Diffuser Actor~\cite{3d_diffuser_actor} leverages 3D point clouds to boost generalization across camera viewpoints. SAM2Act~\cite{fang2025sam2act} improves generalization to diverse scene perturbations by incorporating a memory bank.

For embodiment generalization, Cogact~\cite{li2024cogact}, UniVLA~\cite{bu2025univla}, and SpatialVLA~\cite{qu2025spatialvla} provide examples of good generalization to unseen embodiments and tasks. Hamster~\cite{li2025hamster}, TraceVLA~\cite{zheng2024tracevla}, and OpenVLA~\cite{kim2024openvla} exhibit strong generalization to different embodiments and scenes, accommodating varying lighting and distractors.
Despite these initial approaches, generalization across different embodiments is still limited to similar morphologies and application scenarios. Currently, the generalization of manipulation policies between embodiments with different configurations, particularly those with varying degrees of freedom, has not yet been achieved.

For \textbf{inter-task generalization}, Gato~\cite{reed2022generalist}, PaLM-E~\cite{driess2023palm}, GeminiRob~\cite{team2025gemini}, and 3D-VLA~\cite{zhen20243d} present generalist models that go beyond manipulation tasks. RT-1~\cite{brohan2022rt}, Octo~\cite{team2024octo}, $\ pi_0$~\cite {black2410pi0}, RDT-1B~\cite {liu2024rdt}, GO-1~\cite{bu2025agibot}, and GraspVLA~\cite{deng2025graspvla} improve task generalization by applying scaling laws: more data and model parameters. HiveFm~\cite{guhur2023instruction} and SAM-E~\cite{zhang2024sam} achieve generalization for new RLBench~\cite{james2020rlbench} tasks. ReKep~\cite{huangrekep} and RoboVLM~\cite{li2024towards} show generalization to unseen objects and tasks. DiVLA~\cite{wen2024diffusion} and HybridVLA~\cite{liu2025hybridvla} combine the continuous nature of diffusion-based actions with the reasoning capabilities of autoregressive generation, leading to strong generalization. AffordDP~\cite{wu2024afforddp} enhances generalization by incorporating affordance prediction from VLMs, a widely adopted approach for improving generalization. DreamGen~\cite{jang2025dreamgen} improves generalization by generating robot data using video world models. LAPA~\cite{ye2024latent}, VPDD~\cite{he2025learning}, and GR00T N1~\cite{bjorck2025gr00t} were pretrained with massive action-less data, further enhancing generalization across tasks. Magma~\cite{yang2025magma} develops spatial-temporal intelligence to achieve generalization in long-horizon tasks. VIDEOSAUR~\cite{chapin2025object} reveal that object-centric representations benefit generalization ability. QueST~\cite{mete2025quest} demonstrates generalization to new skills. ADPro~\cite{li2025adpro} proposes a test-time adaptive policy via manifold and initial noise constraints. HiRobot~\cite{shi2025hi}, ECoT~\cite{zawalskirobotic}, and CoT-VLA~\cite{zhao2025cot} utilize the planning and reasoning capabilities of LLM-based CoT to generalize across new tasks and long-horizon challenges.

\subsection{Policy for Data Efficiency}\label{sec:dataefficiency}


In imitation learning, acquiring expert data with action labels is both time-consuming and labor-intensive, making it impractical to prepare expert data for every task or scenario. Given these real-world challenges, research on achieving effective model learning with fewer labeled data—i.e., improving data efficiency—holds significant value.

Several approaches tackle this issue by reducing reliance on labeled data. For instance, PerAct~\cite{shridhar2023peract}, PolarNet~\cite{chen2023polarnet}, DP3~\cite{ze20243d}, and BridgeVLA~\cite{li2025bridgevla} utilize 3D data representations, reducing the need for 2D data from multiple viewpoints. MDT~\cite{reuss2024multimodal}, LAPA~\cite{ye2024latent}, and Hamster~\cite{li2025hamster} are trained using actionless data, minimizing the need for labor-intensive labeling. Equidiff~\cite{wang2024equivariant}, EquiBot~\cite{yangequibot}, ActionFlow~\cite{funk2024actionflow}, and EquAct~\cite{zhu2025equact} design networks that ensure model predictions adhere to equivariance properties, such as rotation, translation, and scaling, which decreases the demand for extensive labeled data. Additionally, other methods like LDuS~\cite{kim2024llm} and ADPro~\cite{li2025adpro} incorporate prior guidance during test time as constraints, avoiding dependence on target task labels.

\subsection{Policy for Sampling Efficiency}\label{sec:samplingefficiency}

Sampling efficiency is critical for high-frequency control and minimizing latency, allowing robots to complete tasks fast and smoothly. This is especially important in dynamic scenes, where rapid reactions are essential for task success.

Early works such as Act3D~\cite{gervet2023act3d} and RVT~\cite{goyal2023rvt} improved action sampling efficiency by replacing information-dense multi-view 2D images with sparse 3D point clouds, reducing computational overhead. Later, methods like ActionFlow~\cite{funk2024actionflow} and FlowMS~\cite{rouxel2024flow} boosted generation efficiency by replacing SDE-based diffusion models with ODE-based flow matching models, which require fewer iterative steps. FMP~\cite{zhang2024affordance} applies flow matching to transform random waypoints into desired action trajectories, while $\pi_0$~\cite{black2410pi0} and OpenVLAOFT\cite{kim2025fine} use flow matching to generate action tokens for faster decoding. SmolVLA~\cite{shukor2025smolvla} employs a 10-step flow matching expert to produce action trunks deployable on consumer-grade GPUs or even CPUs. CotPolicy~\cite{sochopoulos2025fast} further reduces inference cost by using conditional optimal transport to enforce straight solutions in the flow ODE for action generation.

Other works improve efficiency by replacing single-action generation with action trunk, such as ACT~\cite{zhaolearning2023}, VQ-BeT~\cite{lee2024behavior}, and ARP~\cite{zhang2025autoregressive}. Transformer-based methods also benefit from more efficient action token usage: Fast~\cite{pertsch2025fast} introduces a frequency-space action sequence tokenization that greatly compresses token count, while VLACache~\cite{xu2025vla} selects and reuses tokens with minimal changes across steps. CARP~\cite{gong2024carp} adopts a coarse-to-fine autoregressive policy, first learning multi-scale action representations and then refining them via a GPT-style transformer. DiVLA~\cite{wen2024diffusion} and RDP~\cite{xue2025reactive} enhance efficiency by generating shared action tokens in latent space.

We should view the concepts of generalization and data efficiency discussed in each paper from a developmental perspective, as their meanings evolve. Generalization progresses from basic forms to more advanced ones that better align with real-world requirements. Similarly, the expectations for data efficiency evolve with time as technology advances. 

\section{Pretraining methods}\label{sec:Pretraining}
Large-scale pretraining of models has emerged in recent years as a powerful paradigm in natural language processing and computer vision, enabling remarkable advancements. Leveraging vast text corpora, LLMs have transformed text generation and inspired similar breakthroughs in vision through multimodal training. Building on this momentum, models trained on paired image-text data have been introduced to connect visual and linguistic modalities, leading to the development of VLMs. Encouraged by these successes, robotics researchers began to explore whether large-scale models could be extended to embodied domains, where reasoning must be grounded in action and interaction with the physical world. 

In this section, we separate two pretraining methods: pretraining with available action data, and action-free datasets. Each method can include pretraining on \textbf{in-domain} (robotic data containing examples of tasks to be achieved) and/or on \textbf{out-of-domain} data (non-robotic data such as egocentric videos from different tasks and environments). 

\subsection{Pretraining with action data}\label{subsec:withaction}
Reproducing such breakthroughs in robotics, however, has been challenging for a long time due to the scarcity of large-scale action datasets and the inherent difficulty of collecting diverse demonstrations across tasks, environments and embodiments. Early attempts such as Gato \cite{reed2022generalist} and VIMA \cite{jiang2022vima} demonstrated the feasibility of scaling transformers to multi-task, multi-embodiment data. RT-1 \cite{brohan2022rt} and later iterations in the RT-X series were the first to showcase how large-scale real-world robot data could produce scalable, generalizable policies. Other works such as PaLM-E \cite{driess2023palm}, ACT \cite{zhaolearning2023} and RoboFlamingo \cite{li2023vision} began to bridge multimodal reasoning with action execution, highlighting the synergy between language, perception and control. PaLM-E is one of the first VLAs, e.g., using a VLM to predict robotic actions. 

A major step forward came with the Open-X Embodiment dataset~\cite{o2024open}, which pooled a large-scale set of demonstrations across multiple labs, robots and institutions. This large-scale collaborative effort not only expanded the quantity of available trajectories but also introduced unprecedented diversity in embodiments, environments, and task types. 
The Open-X initiative laid the foundation for models such as OpenVLA \cite{kim2024openvla} and 3D-VLA \cite{zhen20243d}, which demonstrated how pre-training on diverse action data can unlock generalization, key ingredients for building general-purpose agents. 

Following the introduction of Open-X, several works have demonstrated that pretraining on large-scale action data is a key enabler of generalization and efficiency in robotics. GR00T-N1 \cite{bjorck2025gr00t} showed that pretraining a VLA model on a heterogeneous mix of real robot demonstrations, human videos and synthetic data equips humanoid robots with transferable sensorimotor priors, yielding robust zero-shot manipulation and faster adaptation compared to training from scratch. Similarly, Physical Intelligence’s $\pi_0$~\cite{black2410pi0} explicitly validated the benefits of action pretraining by layering a flow-matching action head onto a pretrained VLM, allowing the model to leverage broad multi-embodiment datasets for robust generalization. Their follow-up, FAST \cite{pertsch2025fast}, introduced an efficient action tokenization method that preserves the benefits of large-scale pretraining while reducing compute overhead, demonstrating that pretrained action representations can rival diffusion-based policies with greater efficiency. Open-source initiatives have echoed this trend: Octo \cite{team2024octo} highlighted how large-scale action pretraining allows a single model to transfer across diverse observation and action spaces; while SmolVLA \cite{shukor2025smolvla} pushed in the opposite direction, showing that even lightweight models pretrained on community-collected action datasets can match the performance of larger VLAs. Collectively, these efforts illustrate that pretraining on diverse action datasets is not just beneficial but essential, serving as the robotics analogue of LLM pretraining, laying the groundwork for scalable, general-purpose embodied intelligence.

\subsection{Pretraining with action-free data}\label{subsec:action-free}
Despite the advances in robotic pretraining, the scale of publicly available robotic action data remains orders of magnitude smaller than text or image corpora used in large language or vision models. This limitation has spurred the development of emerging approaches that explore pretraining strategies leveraging passive observation, video prediction, or world models to learn sensorimotor priors. 

Notable examples include GR-1 \cite{wuunleashing}, which employs large-scale video generative pretraining to learn multi-task visual robot manipulation without requiring extensive action-labeled data; LAPA \cite{ye2024latent}, which introduces an unsupervised method for pretraining VLA models from internet-scale videos and UniVLA \cite{bu2025univla}, which derives task-centric latent actions from videos, enabling cross-embodiment policy learning without action annotations. Here, leveraging out-of-domain video data can provide large-scale coverage for generalizable priors, while in-domain video data improves task-specific adaptation. These approaches demonstrate that leveraging large-scale, unlabeled video data can effectively pretrain models for robotic tasks, reducing the dependency on action-labeled datasets. 

Another complementary direction focuses on learning compact action representations through discrete latent variable models. Methods such as QueST \cite{mete2025quest} map continuous action sequences to discrete skill tokens using an autoencoder architecture that captures temporal correlations, which are then used to train policies via next-token prediction. VQ-BeT \cite{lee2024behavior} similarly encodes action sequences into discrete latent vectors, learning shared tokens over low-level skills without explicitly modeling temporal correlations. These discrete latent representations enable robust transfer across tasks and embodiments while reducing reliance on large-scale action datasets. 


These action-free pretraining approaches suggest robots can acquire transferable skills in domains with limited action-labeled data, advancing generalist robotic systems.

\section{Evaluations}\label{sec:evaluation}
For the evaluation of each method, we first reviewed each paper, summarizing the core tools or priors, strengths, and limitations, and providing a subjective qualitative assessment. Then, we calculated the average monthly citation count for each method, which reflects, to some extent, the impact of the work. The statistical results are shown in Tab.~\ref{tab:procons}. 

Furthermore, we conducted a comprehensive search and organized qualitative comparison results across multiple benchmarks, providing an objective qualitative evaluation. Below, we will present the datasets, evaluation metrics, and quantitative comparison results related to these qualitative analyses.

\vspace{-2mm}
\subsection{Datasets} 
As shown in Tab.~\ref{tab:taxonomy}, RMP commonly consume 3 observation modalities: i) 1D states (proprioception, joint/EE pose, and compact object states), ii) 2D visual streams (RGB from one or multiple views), and iii) 3D observations (RGB-D or point clouds). To study manipulation and close the sim-to-real gap, the community relies on high-fidelity simulators/environments together with real-robot datasets that mirror these setups. Representative resources include panda-gym~\cite{gallouedec2021panda}, RoboVerse~\cite{geng2025roboverse}, and MuJoCo Playground~\cite{zakka2025mujoco}for light-weight experiments in physics simulators, and large-scale, real-world datasets such as Jacquard~\cite{depierre2018jacquard}, RH20T~\cite{fang2023rh20t}, and Open-X Embodiment~\cite{o2024open} that provide demonstrations aligned with common robot hardware and scenes.

The CALVIN~\cite{mees2022calvin} benchmark is built on top of the PyBullet~\cite{coumans2016pybullet} simulator and involves a Franka Panda Robot arm that manipulates the scene. CALVIN consists of 34 tasks and 4 different environments. All environments are equipped with a desk, a sliding door, a drawer, a button that turns on/off an LED, a switch that controls a lightbulb and three different colored blocks (red, blue and pink). These environments differ from each other in the texture of the desk and positions of the objects. CALVIN provides 24 hours of tele-operated unstructured play data, 35\% of which are annotated with language descriptions. Each instruction chain includes five language instructions that need to be executed sequentially.

RLBench~\cite{james2020rlbench} is built atop the CoppelaSim~\cite{rohmer2013v} simulator, where a Franka Panda Robot is used to manipulate the scene. There are four RGB-D cameras available, front, wrist, left shoulder and right shoulder of the robot.

LBERO~\cite{liu2023libero} consists of over 130 language-conditioned manipulation tasks divided into 5 different task suites: LIBERO-Spatial, LIBERO-Goal, LIBERO-Object, LIBERO-90, and LIBERO-Long. Each task suite except for LIBERO-90 consists of 10 different tasks with 50 demonstrations. Each task suite focuses on different challenges of imitation learning: LIBERO-Goal tests on tasks with similar object categories but different goals. LIBERO-Spatial requires policies to adapt to changing spatial arrangements of the same objects. In contrast, LIBERO-Object maintains he layout while changing the objects. LIBERO-90 consists of 90 different tasks in several environments and tasks with various spatial layouts.

Meta-World~\cite{yu2020meta} is a MuJoCo-based~\cite{zakka2025mujoco} benchmark featuring a Sawyer robot and 50 diverse manipulation tasks (e.g., pick-place, door open, drawer open). It provides standard splits for multi-task and meta-RL: MT10/MT50 for multi-task learning and ML10/ML45 for meta-learning, with disjoint train/test task sets to measure cross-task generalization. Observations can be state vectors or RGB; success is typically measured by binary task completion and goal distance thresholds.

RoboSuite~\cite{zhu2020robosuite} builds on MuJoCo and supports Sawyer, Franka, and other arms with standardized tasks such as Lift, Stack, Door, NutAssembly, Wipe. Multiple camera views are provided (front/side/agent/wrist) alongside low-dimensional states. The robomimic\cite{robomimic2021} datasets supply large-scale human teleoperation demonstrations at varying qualities (expert/medium/mixed), enabling imitation and offline RL benchmarks with consistent metrics and baselines.

ManiSkill~\cite{gu2023maniskill2} offers GPU-accelerated simulation for dozens of tabletop and articulated-object tasks (e.g., PickCube, Stack, OpenCabinetDoor/Drawer, Plug/Unplug). It includes large demonstration sets generated via motion planning and teleoperation, along with language/task descriptions for some suites. Standard metrics include success rate, completion time, and pose/placement accuracy; they emphasize contact-rich tasks and generalization across object instances.

RT-1 / Open-X Embodiment ~\cite{brohan2022rt,o2024open} aggregates large-scale real-world teleoperation data from fleets of mobile manipulators performing hundreds of natural-language-labeled tasks (pick/place, open/close, tidy up). RGB observations (sometimes multi-view) and language commands train transformer policies for broad generalization. Benchmarks evaluate
success rates across seen and novel tasks, cross-robot transfer, and robustness in unconstrained homes and offices.

SimplerEnv~\cite{li2024evaluating} is a simulation suite designed to mirror real Google-Robot tabletop tasks so that policy rankings in sim correlate with rankings on the real robot. Concretely, it reproduces tasks such as Pick Coke Can (PCC), Move Near (MN), and Open/Close Drawer (OCD) and evaluates policies under two regimes: Visual Matching (renderings closely match real scenes) and Variant Aggregation (averaging success over many appearance/layout variations).

The COLOSSEUM~\cite{pumacay2024colosseum} is a generalization benchmark (built on RLBench/CoppeliaSim) that selects 20 manipulation tasks and systematically perturbs the environment along 14 axes (e.g., lighting, colors/textures, distractors, object sizes, physical properties, camera pose). The principal metric is the average decrease in performance from the base setting to perturbed settings (lower is better).

\subsection{Metrics}
\textbf{Success rate} denotes the proportion of attempts that successfully complete the task, which measures how often the robot can perform a given task correctly.

\textbf{Task completion time} is the time taken for the robot to finish the task from start to end. Faster completion indicates higher efficiency, which is important in practical scenarios. 

\textbf{Average length(Avg. Len)} in CALVIN’s long-horizon evaluation, each episode is a chain of $K$ atomic instructions. For rollout $i$, let $L_i$ be the length of the \emph{longest correct prefix}—the number of consecutive sub-tasks completed \emph{in order} before the first error. The metric is $\mathrm{Avg. Len}=\frac{1}{M}\sum_{i=1}^{M} L_i$ computed over $M$ test episodes. It captures partial progress on instruction chains (larger is better) and complements binary success. 

\textbf{Success weighted by Path Length(SPL)} couples \emph{whether} an agent succeeds with \emph{how directly} it moves and is bounded 

\begin{onecolumn}{
\centering
\scriptsize
\linespread{0.99} \selectfont
\begin{longtable}{p{1.8cm}|p{3cm}p{5.5cm}p{5cm}c}    
    \caption{Core Priors or Tools, strengths, and limitations of all selected robotic manipulation methods. AMC denotes average monthly citations.}\\
    \bottomrule 
    \multicolumn{1}{c}{\textbf{Article}} & \multicolumn{1}{c}{\textbf{Core Priors / Tools}} & \multicolumn{1}{c}{\textbf{Strengths}}  &  \multicolumn{1}{c}{\textbf{Limitations}} & \textbf{AMC} \label{tab:procons}\\
    \hline 
    \endhead     
    DP~\cite{chi2023diffusion}&\cellcolor{ImportantColor}time-series diffusion &multimodal and high-dimensional output& low inference efficiency  & 54.2\\
    SuSIE~\cite{black2023zero}&\cellcolor{ImportantColor} pretrained generative model  &  zero-shot generalization to new objects &image generation and  policy trained separately  & 7.2\\
    ChaDiffuser~\cite{xian2023chaineddiffuser}&\cellcolor{ImportantColor} keypose  prediction &improved long-horizon task performance  & keypose prediction errors propagate & 18.12\\
    VPDD~\cite{he2025learning}&\cellcolor{ImportantColor}discrete diffusion model & actionless pretraining from video &lack language support and fine-grained control   & 0.5\\
    DP3~\cite{ze20243d}&\cellcolor{ImportantColor}compact 3D representation & generalize across position, instance, appearance& lack language input support & 10.2\\
    DiffuserActor~\cite{3d_diffuser_actor}&\cellcolor{ImportantColor} 3D scene representations &  robust to scene changes & requires camera calibration, low efficiency & 8.5\\
    Equidiff~\cite{wang2024equivariant}&\cellcolor{ImportantColor} $\SO$(2)-equivariant & sample-efficient learning &lack language support; low sampling efficiency& 2.66\\
    EquiBot~\cite{yangequibot}&\cellcolor{ImportantColor}SIM(3)-equivariant &  sample-efficient learning &lack language support; low sampling efficiency  & 3.23\\
    MDT~\cite{reuss2024multimodal}& \cellcolor{ImportantColor}  goal-image conditioned &excellent performance on CALVIN and LIBERO  &high computational and memory overhead   & 5.16\\
    SDP~\cite{wangsparse}&\cellcolor{ImportantColor}  task-specific routers& skill reuse across tasks & needs task IDs; MoE routing complexity & 1.75\\
    PSEC~\cite{liu2025skill}&\cellcolor{ImportantColor}compositional policies &continual policy shift and dynamic shift settings  & redundant skill expansion & 0.5\\
    AdaManip~\cite{wangadamanip}&\cellcolor{ImportantColor}adaptive data collection & focus on articulated object manipulation &limited to 9 object categories and 5 mechanisms  & 1.25\\
    AffordDP~\cite{wu2024afforddp}&\cellcolor{ImportantColor}transferable affordance & generalize to unseen instances and categories & heavily rely on affordance's accuracy & 0.83\\
    KStar~\cite{lv2025spatial}&\cellcolor{ImportantColor}kinematics constraint & reliable and kinematics-aware action & no significant improvement in scores & 0.67\\
    BRS~\cite{jiang2025behavior}& \cellcolor{ImportantColor}rich user feedback  & combine arms, base, torso for real-world manipulation & limited generalizability & 2.25\\
    CotPolicy~\cite{sochopoulos2025fast}&\cellcolor{ImportantColor}optimal transport~(OT) & competitive success rates with high efficiency &requires rich and aligned demonstrations for OT & 0.0\\
    \hline
    Octo~\cite{team2024octo}&\cellcolor{ImportantColor}scaling law &a versatile policy initialization across platforms  & unstable in long-horizon tasks & 26.69\\
    DiVLA~\cite{wen2024diffusion}&\cellcolor{ImportantColor} contextual reasoning & adaptability to novel instructions and environments & heavy computational cost  & 3.79\\
    Cogact~\cite{li2024cogact}& \cellcolor{ImportantColor} separates cognitive and action  &  study design of action modules and scaling behaviors & limited generalizability & 6.53\\
    ChatVLA~\cite{zhou2025chatvla}&\cellcolor{ImportantColor}mixture of experts & perform well on zero-shot and few-shot tasks & heavy dependence on VLM/LLM accuracy & 4.0\\
    RDT-1B~\cite{liu2024rdt}&\cellcolor{ImportantColor}unified action space & advancements in dexterous bimanual manipulation & heavy computational cost & 13.86\\
    GO-1~\cite{bu2025agibot}& \cellcolor{ImportantColor} latent action model + DP & generalization and dexterity improve with dataset size &low inference efficiency  & 4.61\\
    GR00TN1~\cite{bjorck2025gr00t}&\cellcolor{ImportantColor} latent actions & open-source, cross‑embodiment foundation model &  focus on short-horizon tabletop manipulation tasks & 13.94\\
    DreamGen~\cite{jang2025dreamgen}& \cellcolor{ImportantColor}pseudo-action extraction & learn from actionless videos, broader generalization &heavy computation; limited dexterous behaviors & 0.63\\
    HybridVLA~\cite{liu2025hybridvla}&\cellcolor{ImportantColor}autoregressive+diffusion &robust to unseen objects, layouts, and lighting &low inference efficiency  & 5.15\\
    \hline
    FMP~\cite{zhang2024affordance}&\cellcolor{ImportantColor} affordance prediction &  stable training; faster inference than DP & not robust to scene and camera view & 1.56\\
    RDP~\cite{xue2025reactive}&\cellcolor{ImportantColor}real-time tactile response & improved contact-aware behavior & limited scalability to long-horizon tasks & 2.73\\
    $\pi_0$~\cite{black2410pi0}&\cellcolor{ImportantColor}VLMs; scaling law & generalist robot policies & low sampling efficiency, sensitivity to scene &26.43 \\
    GraspVLA~\cite{deng2025graspvla}& \cellcolor{ImportantColor} 2D bounding boxes&  actionless pretraining from video & only for rigid object grasping & 4.3\\
    HiRobot~\cite{shi2025hi}& \cellcolor{ImportantColor}intermediate text command  &  evaluate on   single-arm, dual-arm, and mobile robots & high-level policy not aware of low-level execution & 7.12\\
    SmolVLA~\cite{shukor2025smolvla}&\cellcolor{ImportantColor}Asynchronous inference  & matches 10× larger VLAs in performance & only for simple and short-horizon tasks & 1.1\\
    \hline
    MVP~\cite{xiao2022masked}& \cellcolor{ImportantColor} masked visual pre-training & without relying on labels or expert demonstrations & policy head must be re-optimized for each task & 6.86\\
    RT-1 \cite{brohan2022rt}&\cellcolor{ImportantColor}pre-trained FiLM-EfficientNet &real-world deployment ready  & lacks closed-loop feedback; discrete action& 41.92\\
    RoboUniView~\cite{liu2024robouniview}&\cellcolor{ImportantColor}vision-language alignment &generalization across various camera parameters  &rely on precise camera calibration  & 0.83\\
    Lift3D~\cite{jia2024lift3d}&\cellcolor{ImportantColor}task-related affordance mask &  enhanced 3D spatial awareness & no language control & 2.38\\
    SAM2Act~\cite{fang2025sam2act}& \cellcolor{ImportantColor} memory bank; SAM+LoRA &  a strong understanding of spatial memory & maintaining memory in new scenes is challengings& 1.04\\
    OpenVLAOFT~\cite{kim2025fine}& \cellcolor{ImportantColor}  parallel decode+action chunk & enhanced  efficiency, policy performance and flexibility & struggle to model multimodal action distributions  & 7.67\\
    \hline
    Gato~\cite{reed2022generalist}&\cellcolor{ImportantColor}autoregressive Transformers  & a multi-task multi-embodiment generalist policy & limited task performance & 31.31\\
    VIMA~\cite{jiang2022vima}& \cellcolor{ImportantColor}latent action model  & object-centric tokens enable scalable learning & high computational and memory overhead & 8.4\\
    PaLM-E~\cite{driess2023palm}&\cellcolor{ImportantColor}chain-of-thought reasoning & scalable and general-purpose & heavy resource requirements & 133.3\\
    ACT~\cite{zhaolearning2023}&\cellcolor{ImportantColor}chunking with Transformer  &low-cost hardware enables high-precision operation  & chunk length sensitivity; limited generalization & 25.76\\
    GR-1~\cite{wuunleashing}&\cellcolor{ImportantColor} pretrained generative model& foundation VLA model & limited generalization, discrete actions & 6.0\\
    RoboFlamingo~\cite{li2023vision}&\cellcolor{ImportantColor}pre-trained VLMs & enhanced performance and generalization &low computational efficiency  & 9.25\\
    3D-VLA~\cite{zhen20243d}&\cellcolor{ImportantColor} object and location tokens & 3D-aware world modeling & limited closed-loop feedback & 7.17\\
    VQ-BeT~\cite{lee2024behavior}&\cellcolor{ImportantColor}common latent action tokens  & sampling efficiency; multimodal action prediction &lack language input support  & 5.1\\
    OpenVLA~\cite{kim2024openvla}& \cellcolor{ImportantColor}  VLM + diverse dataset & outperform RT-2-X by 16.5\% across 29 tasks  & low inference throughput and robustness & 47.95\\
    QueST~\cite{mete2025quest}&\cellcolor{ImportantColor}common latent skill tokens &  superior long-horizon modeling capabilities & no scene/object awareness in skill tokens& 1.27\\
    ARP~\cite{zhang2025autoregressive}&\cellcolor{ImportantColor}causal chunked Transformer & robustness across various control frequencies &need multiple high-quality expert demonstrations  & 1.16\\
    CARP~\cite{gong2024carp}&\cellcolor{ImportantColor}coarse-to-fine refinement & competitive success rates with high efficiency &lack language support and losed-loop feedback  & 0.95\\
    TraceVLA~\cite{zheng2024tracevla}& \cellcolor{ImportantColor} visual trace prompting & robust generalization across embodiments and scenarios & high computational and memory overhead  & 4.26\\
    RoboVLM~\cite{li2024towards}& \cellcolor{ImportantColor} VLMs+history fusion  & detailed guidebook for the design of VLAs & limited long-horizon, complex task ability & 3.85\\
    Fast~\cite{pertsch2025fast}&\cellcolor{ImportantColor}discrete cosine transform & good at dexterous and high-frequency tasks &low inference efficiency  & 10.00\\
    SpatialVLA~\cite{qu2025spatialvla}& \cellcolor{ImportantColor} adaptive action grids & strong zero-shot performance, high-frequency control &action discretization limit precision  & 5.34\\
    VLACache~\cite{xu2025vla}&\cellcolor{ImportantColor}reuse unchanged tokens &achieve 1.7× faster with comparable performance& cache effectiveness depends on visual stability & 1.49\\
    Hamster~\cite{li2025hamster}&\cellcolor{ImportantColor} 2D path guidance & robust spatially-aware action generation & lack spatial 3D understanding & 3.25\\
    Magma~\cite{yang2025magma}& \cellcolor{ImportantColor}set-of-mark;  trace-of-mark&  enhanced the spatial-temporal intelligence  & high computational and memory overhead   & 5.28\\
    CoT-VLA~\cite{zhao2025cot}& \cellcolor{ImportantColor} visual chain-of-thought  &  outperforming OpenVLA by 17\% in real-world  tasks  & high computational and memory overhead & 9.12\\
    UniVLA~\cite{bu2025univla}&\cellcolor{ImportantColor} latent action learning &  unified, embodiment-agnostic action space & action discretization limit precision & 1.0\\
    LAPA~\cite{ye2024latent}& \cellcolor{ImportantColor}latent action via visual change &unsupervised pretraining from actionless videos & frame-level focus limits long-horizon reasoning & 5.41\\
    ECoT~\cite{zawalskirobotic}&\cellcolor{ImportantColor}chain-of-thought reasoning& generalization to novel tasks & data and prompt engineering sensitivity & 7.28\\
    WorldVLA~\cite{cen2025worldvla}& \cellcolor{ImportantColor} attention mask strategy& closed-loop action and observation prediction  & action discretization limit precision & 0.0\\
    \hline
    LOTUS~\cite{wan2024lotus}&\cellcolor{ImportantColor} continual skill discovery&  unsupervised skills from raw, unsegmented demos& replay buffer grows with tasks & 1.49\\
    HULC~\cite{mees2022hulc}&\cellcolor{ImportantColor}visual-language alignment &systematic analysis of impact of various factors & low success rate& 4.45\\
    BridgeVLA~\cite{li2025bridgevla}&\cellcolor{ImportantColor}2D heatmaps  & real‑robot success with high sample efficiency & rely on precise camera calibration  & 0.0\\
    SEDF~\cite{urain2023se}&\cellcolor{ImportantColor}$\SE$(3)-equivariance & sample-efficient learning & only object-level grasping & 18.1\\
    A0~\cite{xu2025a0}& \cellcolor{ImportantColor}affordance representation & easy to deploy across different robotic platforms &rely on depth and camera calibration accuracy  & 2.5\\
    FlowMS~\cite{rouxel2024flow}&\cellcolor{ImportantColor} smooth action vector field & whole-body movements on a full-size humanoid robot &success rate and diversity lower than DPs.  & 1.27\\
    ActionFlow~\cite{funk2024actionflow}&\cellcolor{ImportantColor}$\SE$(3)-equivariance & high efficiency and low latency & assumes smooth transitions & 0.11\\
    PolarNet~\cite{chen2023polarnet}& \cellcolor{ImportantColor}3D PC representation & outperforms SOTA in single- and multi-task settings & limited generalization to new scenes, objects, tasks & 2.1\\
    HiveFm~\cite{guhur2023instruction}& \cellcolor{ImportantColor}history-aware Transformer & outperform SOTA baselines on 74 RLBench tasks & quadratic computational cost with sequence length & 3.78\\
    \hline
    PerAct~\cite{shridhar2023peract}&\cellcolor{ImportantColor} voxel-based formulation&  data efficiency & slow inference speed, discretized actions & 16.87\\
    RVT~\cite{goyal2023rvt}&\cellcolor{ImportantColor}multi-view fusion &enhanced geometric reasoning  &sensitive to camera configuration  & 6.5\\
    Act3D~\cite{gervet2023act3d}& \cellcolor{ImportantColor}relative 3D cross-attentions  &generalize well to novel camera placements  & rely on precise camera calibration & 3.45\\
    SAM-E~\cite{zhang2024sam}& \cellcolor{ImportantColor}SAM+LoRA & improve generalization to new tasks  &performance is sensitive to segmentation quality  & 1.09\\
    EquAct~\cite{zhu2025equact}&\cellcolor{ImportantColor}$\SE$(3)-equivariance & generalize better across object poses and orientations &low sampling efficiency & 0.0\\
    \hline
    Cliport~\cite{shridhar2022cliport} &\cellcolor{ImportantColor} affordance prediction & with spatial reasoning & discrete actions; just for tabletop tasks & 13.09\\
    MOKA~\cite{liu2024moka}& \cellcolor{ImportantColor}affordance in-context learning & enable zero/few-shot open-world manipulation & high latency; limited affordance accuracy & 4.2\\
    RAM~\cite{kuang2024ram}& \cellcolor{ImportantColor} retrieval-based affordance&  generalizable zero-shot robotic manipulation  &struggle with long-horizon and complex actions  & 2.6\\
    ReKep~\cite{huangrekep}&\cellcolor{ImportantColor}keypoint constraints & consider spatio-temporal dependencies & rely on accurate point tracking & 12.11\\
    3D-LOTUS~\cite{garcia2024towards}&\cellcolor{ImportantColor}pretrained VLMs & strong few-shot performance & over-reliance on VLM for planning and grounding & 1.12\\
    GeminiRob~\cite{team2025gemini}&\cellcolor{ImportantColor}Gemini 2.0;  vast dataset & generalist model capable of directly controlling robots& struggle with grounding spatial relationships & 7.0\\
    GeminiRob~\cite{team2025gemini}&\cellcolor{ImportantColor}object-centric interaction & robust control without requiring VLM fine-tuning &reliance on high-quality object meshes  &2.32 \\
    \hline
    \end{longtable}%
    
}
\end{onecolumn}
\twocolumn

\noindent in $[0,1]$. For episode $i$, let $S_i\!\in\!\{0,1\}$ indicate success, $\ell_i^*$ be the shortest-path length (e.g., geodesic) from start to goal, and $\ell_i$ the agent’s executed path length.
The dataset-level score is
\begin{equation}
    \mathrm{SPL}=\frac{1}{M}\sum_{i=1}^{M} S_i\,\frac{\ell_i^*}{\max(\ell_i,\;\ell_i^*)}
\end{equation}

For long-horizon manipulation (e.g., Franka Kitchen~\cite{fu2020d4rl}), papers report \textbf{sub-goals achieved} (microwave open, burner on, door open, etc.) in addition to binary success. 

\textbf{Generalization} quantifies how performance changes under controlled perturbations. The COLOSSEUM~\cite{pumacay2024colosseum} perturbs lighting, textures, distractors, camera, object properties, etc., and reports the \textbf{average performance decrease} from the base setting across perturbation axes ($\downarrow$ is better). SimplerEnv~\cite{li2024evaluating} reports \textbf{MMRV} (Mean Maximum Rank Violation, $\downarrow$) and \textbf{Pearson} ($\downarrow$) to measure sim-to-real rank correlation, alongside success rates under “visual matching” and “variant aggregation.” These metrics explicitly target robustness and sim-to-real alignment rather than raw success alone

\subsection{Qualitative evaluations} 
To provide a fair and comprehensive evaluation of existing RMPs, we first searched through publicly available results and got the leaderboards of CALVIN and RLBench. On the other hand, we reviewed the qualitative comparison results presented in the papers and recorded the outcomes obtained under the same benchmarks and experimental setups. Here, we present the results from benchmarks with six or more comparison methods, specifically focusing on LIBERO, MetaWorld, SimplerEnv-Google Robot, and COLOSSEUM. These qualitative comparisons evaluate existing RMPs from different perspectives, using various task setups and metrics.

\begin{table}[htb!]
\caption{Zero-shot (Train A, B, C $\rightarrow$Test D) long-horizon evaluation on CALVIN. Avg. Len means average length of the correct trajectory.}
\footnotesize
\renewcommand{\arraystretch}{0.9}
\centering
\begin{tabular}{@{}l|ccccc|c@{}}
\toprule
\multirow{2}{*}{Method}  & \multicolumn{5}{c}{Task completed in a row}& \multirow{2}{*}{Avg. Len $\uparrow$}  \\
& 1 & 2 & 3 & 4 & 5&\\
    \midrule
    MCIL~\cite{lynch2020language}  & 30.4 & 1.3 & 0.2 & 0.0 & 0.0 & \cellcolor{ImportantColor}0.31 \\
    HULC~\cite{mees2022hulc}  & 41.8 & 16.5 & 5.7 & 1.9 & 1.1 & \cellcolor{ImportantColor}0.67 \\
    ChaDiffuser~\cite{xian2023chaineddiffuser} & 49.9 & 21.1 & 8.0 & 3.5 & 1.5 &\cellcolor{ImportantColor} 0.84 \\
    SPIL~\cite{zhou2024language} & 74.2 & 46.3 & 27.6 & 14.7 & 8.0 &\cellcolor{ImportantColor} 1.71 \\
    RoboFlamingo~\cite{li2023vision}  & 82.4 & 61.9 & 46.6 & 33.1 & 23.5 & \cellcolor{ImportantColor}2.48 \\
SIE~\cite{black2023zero}  & 87.0 & 69.0 & 49.0 & 38.0 & 26.0 &\cellcolor{ImportantColor} 2.69 \\
    DeeR~\cite{black2023zero}  & 86.2 & 70.1 & 51.8 & 41.5 & 30.4 &\cellcolor{ImportantColor} 2.82 \\
    GR-1~\cite{wuunleashing} & 85.4 & 71.2 & 59.6 & 49.7 & 40.1 &\cellcolor{ImportantColor} 3.06 \\
    DP3~\cite{ze20243d}  & 53.9 & 44.7 & 38.0 & 34.3 & 29.0 & \cellcolor{ImportantColor}2.00  \\
    Diffuser~\cite{3d_diffuser_actor} & 93.8 & 80.3 & 66.2 & 53.3 & 41.2 & \cellcolor{ImportantColor}3.35 \\
    CLOVER~\cite{yang2024closed} & 96.0 & 83.5 & 70.8 & 57.5 & 45.4 & \cellcolor{ImportantColor}3.53 \\
    DTP~\cite{hou2024diffusion} & 94.5 & 82.5 & 72.8 & 61.3 & 50.0 & \cellcolor{ImportantColor}3.61 \\
    RoboUniView~\cite{liu2024robouniview} & 94.2 & 84.2 & 73.4 & 62.2 & 50.7 & \cellcolor{ImportantColor}3.64 \\
    ADPro~\cite{li2025adpro} & 94.7 & 83.0 & 73.6 & 61.4 & 51.1 & \cellcolor{ImportantColor}3.64 \\
    GHIL-Glue~\cite{hatch2024ghil} & 95.2 & 88.5 & 73.2 & 62.5 & 49.8 & \cellcolor{ImportantColor}3.69 \\
    UniVLA~\cite{bu2025univla} & 95.5 & 85.8  & 75.4 & 66.9 & 56.5 & \cellcolor{ImportantColor}3.80 \\
    MoDE~\cite{reuss2024efficient} & 96.2 & 88.9 & 81.1 & 71.8 & 63.5 & \cellcolor{ImportantColor}4.01 \\
    GR-MG~\cite{li2025gr}  & 96.8 & 89.3 & 81.5 & 72.7 & 64.4 & \cellcolor{ImportantColor}4.04 \\
    RoboVLM~\cite{li2024towards}& 98.0 & 93.6 & 85.4 & 77.8 & 70.4 & \cellcolor{ImportantColor}4.25 \\
    SeeR-Large~\cite{tian2025predictive}  & 96.3 & 91.6 & 86.1 & 80.3 & 74.0 & \cellcolor{ImportantColor}\textbf{4.28} \\

\bottomrule
\end{tabular}
\label{tab:calvin}
\end{table}

Table~\ref{tab:calvin} presents the results from the publicly available leaderboard on CALVIN, which includes 34 tasks across 4 different environments (A, B, C, and D). All methods are evaluated under a zero-shot generalization setup, where models are trained in environments A, B, and C, and tested in environment D. The evaluation metric used is the success rate.

\begin{table}[htb!]
\caption{Average success rate on RLBench (18 tasks, 100 demo/task).}
\footnotesize
\renewcommand{\tabcolsep}{0.9mm}
\centering
\begin{tabular}{lcccc}
\toprule
EquAct~\cite{zhu2025equact} & BridgeVLA~\cite{li2025bridgevla}& SAM2Act~\cite{fang2025sam2act} &ARP~\cite{zhang2025autoregressive}   \\
\rowcolor{ImportantColor} 89.4 &88.2 & 86.8& 84.9\\
\midrule
ADPro~\cite{li2025adpro} & 3D-LOTUS~\cite{garcia2024towards} &RVT-2~\cite{goyal2024rvt}  &Diffuser Actor~\cite{3d_diffuser_actor}\\
83.9  &83.1 &81.4 &81.3 \\
\midrule
Mi-Diffuser~\cite{hu2025train}  &SAM-E~\cite{zhang2024sam}&  Act3D~\cite{gervet2023act3d}& RVT~\cite{goyal2023rvt} \\
\rowcolor{ImportantColor}  77.6 &70.6 &65.0& 62.9 	  \\
\midrule
PerAct~\cite{shridhar2023peract}&  PolarNet~\cite{chen2023polarnet}  &HiveFm~\cite{guhur2023instruction}&  C2F~\cite{james2022coarse}   \\
\rowcolor{ImportantColor}  49.4 &46.4 &45.3 &20.1	  \\
\bottomrule
\end{tabular}
\label{tab:rlbench}
\end{table}

Table~\ref{tab:rlbench} reports the results from the RLBench leaderboard, which includes 18 manipulation test tasks, each with 2 to 60 variations. All methods are trained to predict the next end-effector keypose, and the evaluation metric is task success rate, representing the proportion of execution trajectories that meet the goal conditions specified in language instructions.

\begin{table}[htb!]
\caption{Success rate on LIBERO with four settings (spatial, object, goal, long-horizon). VLA Pt refers pretraining on robotics data. The first part refers to fine-tuning on the entire training set. The second part involves few-shot learning using a small subset of data. The last row uses zero-shot setting.}
\footnotesize
\renewcommand{\arraystretch}{0.8}
\renewcommand{\tabcolsep}{0.9mm}
\centering
\begin{tabular}{lcccccc}
\toprule
Method  & VLA Pt  & Spatial& Object&  Goal &	Long & Avg.\\
\midrule
SmolVLA~\cite{shukor2025smolvla}&\xmark &93 &94 & 91& 77 &\cellcolor{ImportantColor}88.75 \\
WorldVLA~\cite{cen2025worldvla}&\xmark &87.6 &96.2 &83.4 & 60.0  &\cellcolor{ImportantColor}81.8 \\
MDT~\cite{reuss2024multimodal}&\xmark &78.5 & 87.5& 73.5&  64.8    &\cellcolor{ImportantColor} 76.1 \\
DP~\cite{chi2023diffusion}&\xmark &78.3 & 92.5& 68.3&  50.5    &\cellcolor{ImportantColor} 72.4 \\
OpenVLAOFT~\cite{kim2025fine}&\cmark &96.9 & 98.1 & 95.5 & 91.1    &\cellcolor{ImportantColor}95.4  \\
UniVLA~\cite{bu2025univla}&\cmark &96.5 &  96.8 &  95.6 &  92.0    &\cellcolor{ImportantColor} 95.2  \\
$\pi_0$~\cite{black2410pi0}&\cmark &90 &86  & 95 &  73  &\cellcolor{ImportantColor}86.0 \\
MaIL~\cite{jia2024mail}&\cmark &74.3 & 90.1  &  81.8 &  78.6 &\cellcolor{ImportantColor} 83.5  \\
DiT Policy~\cite{hou2024diffusion}&\cmark &84.2 & 96.3 &85.4  &  63.8  &\cellcolor{ImportantColor} 82.4 \\
CoT-VLA~\cite{zhao2025cot}&\cmark &87.5 &  91.6 &87.6  &  69.0  &\cellcolor{ImportantColor} 81.13 \\
OpenVLA~\cite{kim2024openvla}&\cmark &84.7 & 88.4  &79.2  & 53.7  &\cellcolor{ImportantColor} 76.5 \\
Octo~\cite{team2024octo}&\cmark &78.9 &  85.7& 84.6   & 51.1    &\cellcolor{ImportantColor} 75.1 \\
VLACache~\cite{xu2025vla}&\cmark &83.8 &  85.8& 76.4   & 52.8    &\cellcolor{ImportantColor} 74.7 \\
LAPA~\cite{ye2024latent}&\cmark &73.8 & 74.6 & 58.8   & 55.4     &\cellcolor{ImportantColor} 65.7 \\
\midrule
Magma~\cite{yang2025magma}(10 demos)&\cmark &26 & 49& 29 &  -   &\cellcolor{ImportantColor} - \\
OpenVLA~\cite{kim2024openvla}(10 demos)&\cmark &8 & 24&  10 &  -    &\cellcolor{ImportantColor} - \\
\midrule
GraspVLA~\cite{deng2025graspvla}&\cmark &- & 94.1&  91.2 &  82.0    &\cellcolor{ImportantColor} - \\

\bottomrule
\end{tabular}
\label{tab:LIBERO}
\end{table}

Table~\ref{tab:LIBERO} presents the results on LIBERO, as reported in their official paper. Methods are tested on all tasks with 20 rollouts each, and the success rate is averaged over 3 different seeds.

\begin{table}[htb!]
\caption{Average success rate on MetaWorld different difficulty levels.}
\footnotesize
\renewcommand{\arraystretch}{0.9}
\renewcommand{\tabcolsep}{0.9mm}
\centering
\begin{tabular}{lccccc}
\toprule
Method  & Easy & Medium &   Hard  &	VeryHard & Avg.\\
\midrule
Lift3D \cite{jia2024lift3d} &93.1 & 82.4 & 88.0 & 28.0 & \cellcolor{ImportantColor}84.5 \\
SmolVLA~\cite{shukor2025smolvla} &87.14 & 51.82 &  70 & 64  &\cellcolor{ImportantColor}68.24 \\
DP3~\cite{ze20243d} &85.7 &  49.6 &57.0  & 18.0   &\cellcolor{ImportantColor} 65.3  \\
$\pi_0$~\cite{black2410pi0} &80.4 & 40.9 & 36.7  &  44.0   &\cellcolor{ImportantColor}50.5 \\
TinyVLA~\cite{zhou2024tinyllava} &77.6  &21.5 &11.4  & 15.8  &\cellcolor{ImportantColor}31.6\\
DP~\cite{chi2023diffusion} &23.1 & 10.7 & 1.9   &  6.1     &\cellcolor{ImportantColor} 10.5 \\
\bottomrule
\end{tabular}
\label{tab:MetaWorld}
\end{table}

Table~\ref{tab:MetaWorld} reports the average success rate of six RMPs on MetaWorld, a tabletop environment featuring a Sawyer arm with a gripper. The test tasks are divided into four difficulty levels and are captured from two corner camera perspectives. The task categories are as: easy tasks include \textit{button-press, drawer-open, reach, handle-pull, peg-unplug-side, lever-pull, and dial-turn}; medium tasks include \textit{hammer, sweep-into, bin-picking, push-wall, and box-close}; hard and very hard tasks include \textit{assembly, hand-insert, and shelf-place}.

\begin{table}[htb!]
\caption{Success rate on SimplerEnv-Google Robot.}
\footnotesize
\renewcommand{\arraystretch}{0.9}
\renewcommand{\tabcolsep}{0.7mm}
\centering
\begin{tabular}{lcccc|cccc}
\toprule
\multicolumn{1}{c}{\multirow{2}[1]{*}{Method}} &\multicolumn{4}{c}{Visual Matching} &\multicolumn{4}{c}{Variant Aggregation}\\
& PCC &MN & OCD & Avg. & PCC &MN & OCD & Avg.\\
\midrule
SoFar~\cite{qi2025sofar} &92.3 & 91.7 & 40.3 & \cellcolor{ImportantColor}74.9 &90.7 & 74.0 &29.7 &\cellcolor{ImportantColor}67.6\\
Cogact~\cite{li2024cogact} &91.3 & 85.0 & 71.8 & \cellcolor{ImportantColor}74.8 &89.6 & 80.8 & 28.3 & \cellcolor{ImportantColor}61.3\\
VLACache~\cite{xu2025vla} &92.0 & 83.3 &  70.5&\cellcolor{ImportantColor}74.4 &91.7 & 79.3 &  32.5  &\cellcolor{ImportantColor}62.3\\
SpatialVLA~\cite{qu2025spatialvla} & 81.0 & 69.6 &59.3 & \cellcolor{ImportantColor}71.9 &89.5 & 71.7	&36.2 &\cellcolor{ImportantColor}68.8\\
$\pi_0$~\cite{black2410pi0} &88.0 & 80.3& 56.0& \cellcolor{ImportantColor}70.1 &- &- & -& -\\
RoboVLM~\cite{li2024towards} & 72.7 & 66.3& 26.8  & \cellcolor{ImportantColor}56.3 & 68.3& 56.0& 8.5 &\cellcolor{ImportantColor} 46.3\\
RT-1~\cite{brohan2022rt} &56.7& 31.7& 59.7 & \cellcolor{ImportantColor}53.4 & 49.0& 32.3& 29.4  &\cellcolor{ImportantColor}39.6\\
TraceVLA~\cite{zheng2024tracevla} &28.0& 53.7& 57.0 & \cellcolor{ImportantColor}42.0 & 60.0& 56.4& 31.0 &\cellcolor{ImportantColor}45.0\\
OpenVLA~\cite{kim2024openvla} & 16.3 & 46.2 & 35.6  & \cellcolor{ImportantColor} 27.7 & 54.5& 47.7& 17.7  &\cellcolor{ImportantColor}39.8\\
Octo~\cite{team2024octo} & 17.0& 4.2& 22.7 & \cellcolor{ImportantColor}16.8 & 0.6& 3.1& 1.1 &\cellcolor{ImportantColor}1.1\\
\bottomrule
\end{tabular}
\label{tab:SimplerEnv}
\end{table}

Table~\ref{tab:SimplerEnv} presents results on SimplerEnv with Google Robot setups, which offer diverse manipulation scenarios under varying lighting, color, texture, and robot camera pose conditions. In Tab.~\ref{tab:SimplerEnv}, PCC, MN, and OCD denote the tasks \textit{Pick Coke Can}, \textit{Move Near}, and \textit{Open/Close Drawer}.

\begin{table}[htb!]
\caption{Robot Manipulation Generalization on The COLOSSEUM. The metric is the average decrease across all perturbations.}
\footnotesize
\renewcommand{\arraystretch}{0.8}
\renewcommand{\tabcolsep}{0.7mm}
\centering
\begin{tabular}{lcccc}
\toprule
SAM2Act~\cite{fang2025sam2act}& 	
RVT~\cite{goyal2023rvt} & Diffuser~\cite{3d_diffuser_actor} &	
MVP~\cite{xiao2022masked} & PerAct~\cite{shridhar2023peract}\\
\rowcolor{ImportantColor}-4.3 &-14.5 &-15.6& -16.3 & -17.3\\
\midrule
RVT-2~\cite{goyal2024rvt} & GENIMA~\cite{shridhar2024generative} &BridgeVLA~\cite{li2025bridgevla} & R3M~\cite{nair2023r3m}  &	
ACT~\cite{zhaolearning2023} \\
\rowcolor{ImportantColor}-19.5 &-41.6 & -45.3& -49.9& -61.8\\
\bottomrule
\end{tabular}
\label{tab:COLOSSEUM}
\end{table}

Table~\ref{tab:COLOSSEUM} reports the results on the COLOSSEUM benchmark, which extends RLBench. All models are trained using data from the original RLBench but evaluated in environments that span 12 axes of perturbations. These perturbations include variations in object texture, color, size, and changes in background, lighting, distractors, and camera poses. In total, COLOSSEUM generates 20,371 unique task perturbation instances to evaluate model’s generalization capabilities. Specifically, methods are evaluated by performing each task across 25 trials per perturbation, and the average decrease across all perturbations is computed to assess the generalization.

\section{Applications}\label{sec:applications}

Real-world manipulation methods span heterogeneous demands, and we group them into: (i) \emph{Primitive Manipulation Tasks}—short, one-shot skills where perception-to-control dominates; (ii) \emph{Contact-Rich Assembly Tasks}—tight-tolerance insertions, creasing, and continuous-force operations; (iii) \emph{Kitchen Assistant}—long-horizon domestic routines under language or goal guidance; (iv) \emph{Tool-Mediated Manipulation}—skills that require exploiting external affordances; and (v) \emph{Garbage Cleaning}—perception-guided pickup and wiping for litter removal and tabletop cleaning

\subsection{Primitive Manipulation Tasks}
These are short, one-shot manipulations (grasp–place, press, slide, open/close) whose core challenge is precise perception-to-control under noise and contact. Representative methods: GraspVLA~\cite{deng2025graspvla}: open-set tabletop grasping of novel objects; PerAct~\cite{shridhar2023peract}: RLBench primitives like pick/place blocks, rotate knobs, press switches via language; Act3D~\cite{gervet2023act3d}: multi-view 3D grasping, peg/slot insertion, button pressing; RDP~\cite{xue2025reactive}: contact-rich peg-in-hole / connector insertion; CLIPort~\cite{shridhar2022cliport}: pick-and-place and sorting by attributes; RT-1~\cite{brohan2022rt}: hundreds of household primitives like opening doors, tossing trash, moving items to bins. Strong reliability in controlled setups and good breadth with data scaling; still brittle for transparent/deformable items, odd viewpoints, and sim-to-real—needs better tactile fusion and uncertainty-aware control.

\subsection{Contact-rich Assembly Tasks} 
This category targets operations that require sustained contact, precise alignment, and deformation control beyond simple pick-and-place. Manual2Skill~\cite{tie2025manual2skill} parses human manuals into step-wise skills to execute long-horizon furniture assembly with reliable multi-contact sequences. SRSA~\cite{guo2025srsa} focuses on industrial parts assembly, emphasizing stable contact formation and alignment under tight tolerances. The generalist policy $\pi_0$~\cite{black2410pi0} demonstrates cardboard-box folding, a deformable, contact-dominant procedure involving creasing, pressing, and closure. Complementing assembly, RDP~\cite{xue2025reactive} provides reactive control for peeling and wiping, covering surface-preparation and post-assembly finishing where continuous contact and friction management are critical. GarmentPile~\cite{Wu_2025_CVPR} extends contact-rich manipulation to deformable textiles in clutter, learning point-level affordances to retrieve and reorganize piled garments as a precursor to folding or sorting. Together, these methods illustrate progress from isolated grasps toward crease/bend, insert/fit, and surface-treatment behaviors executed robustly on real hardware.

\subsection{Kitchen Assistant}
Robotic policies act as versatile household helpers that fetch items, open/close storage, place utensils/food, and perform light prep/cleanup under language guidance. Representative systems include SayCan~\cite{ahn2022can} and RT-1~\cite{brohan2022rt}. Recent generalist VLA models extend this further: $\pi_0$\cite{black2410pi0} demonstrates broad kitchen skills (e.g., coffee making, dish handling, table setting) as part of its scaled household repertoire, while a liquid-mixing system\cite{khan2025shake} targets beverage preparation with bimanual pouring, quantity control, and step-wise recipe execution. Together, these works show reliable item retrieval, placement, container operations, and simple food/beverage preparation in real kitchens.

\subsection{Tool-Mediated Manipulation}
The goal is to acquire and stabilize a tool and exploit its affordances, demanding functional understanding and contact modeling. Representative methods:  MVP~\cite{xiao2022masked}: simulated levering with a bar and hook-and-pull actions; GR-1/GR-2~\cite{wuunleashing}: evaluations on cutting with a knife and scooping with a spoon/cup after video pretraining; HULC~\cite{mees2022hulc}: everyday wiping/cleaning with a cloth as tool-like behavior; Magma~\cite{yang2025magma}: demonstrations of using kitchen utensils (e.g., spoon scooping, spatula flipping); MOKA~\cite{liu2024moka}: real-robot evaluations used everyday tools and appliances
ToolBench / MuJoCo-Manipulus: benchmark tasks like pouring, scooping, scraping, hammering used by general VLA policies. Clear progress on canonical cases, but robust compositional generalization to unseen tools/targets, liquids and deformables, and varying friction/compliance remains limited; richer contact sensing, dynamics priors, and tool-centric data are needed.

\subsection{Garbage Cleaning}
Garbage cleaning spans aquatic, tabletop, and outdoor settings. Haldorai \etal~\cite{haldorai2024improved} present a vision-guided water-surface robot that detects and collects floating trash in real aquatic environments. GScbam-Net~\cite{li2023learning} offers a lightweight attention-based garbage image classifier to support waste sorting and downstream robotic cleaning. Lv \etal~\cite{lv2023machine} develop ML-based garbage detection and 3D localization to enable autonomous trash pickup in the field.

\section{Discussion and Conclusion} \label{sec:conclusion}

Data-driven robot learning faces a fundamental bottleneck: the scarcity of high-quality and diverse in-domain data. Unlike fields such as natural language processing or computer vision, where massive datasets are readily available, robotic data is costly to collect and often narrow in scope, limiting the generalization ability of learned policies. This calls for \textbf{smarter learning strategies} that embed inductive biases into robotic models, reducing dependence on exhaustive data coverage. As argued in our recent position paper \cite{chen:hal-05086677}, a promising path forward is to design \textbf{bio-inspired foundation models for robotics}, where principles drawn from human and animal perception, motor control, and adaptation can provide the structural priors necessary to scale beyond current limitations.

\subsection{Commonly Used Prior}

In RMP, several forms of prior knowledge are commonly utilized to improve performance. \textbf{Chain-of-thought} (CoT) is one of the most widely used priors~\cite{driess2023palm, zawalskirobotic, zhao2025cot}. CoT, derived from LLMs, enables robots to break down complex or long-horizon tasks into manageable subtasks, facilitating more effective decision-making during manipulation. \textbf{Affordance knowledge}~\cite{wu2024afforddp,liu2024moka,zhang2024affordance} allows robots to recognize and predict how objects can be interacted with, guiding their actions based on the physical properties and potential uses of objects. It is often employed to enhance a model’s generalization ability. Waypoints\cite{zhang2024affordance, mehta2024waypoint} are a specific form of affordance, defining intermediate positions or goals in a manipulation task to help robots plan and execute more precise and efficient movements. \textbf{Kinematic constraints}~\cite{lv2025spatial} ensure that robots respect physical limitations during movement, optimizing their trajectories and minimizing errors. Additionally, \textbf{equivariance constraints}~\cite{wang2024equivariant,yangequibot,funk2024actionflow,zhu2025equact} can be applied to improve the efficiency of policy learning by reducing the required data size. These constraints help the model learn invariant behaviors under specific transformations, thus reducing the demand for data volume. Moreover, other priors, such as \textbf{manifold constraints}~\cite{li2025adpro} derived from observation, are used to fully leverage the information from test-time data. 

\subsection{Summary of open challenges}

\paragraph{\textbf{Insufficient Generalization}}
RMP often struggles with generalization, especially when applied to real-world scenarios. These agents are highly sensitive to the specific scenes or tasks that they are trained on. Their performance deteriorates significantly when confronted with variations in the environment, object types, or task conditions. In particular, the ability to generalize to real-world tasks remains limited, preventing robots from independently handling everyday tasks in dynamic and unstructured settings. 

\paragraph{\textbf{Diverse Robot Configurations}}
The wide variety of robot designs and configurations leads to significant differences in the input and output spaces for learning manipulation policies. These discrepancies make it difficult for existing models to generalize across different robotic platforms. As a result, a unified manipulation policy that can work across various robot types is challenging to achieve. To address this, having a standardized robot configuration is essential for building a versatile and practical foundational VLA model. This would allow for more effective and transferable learning across different robotic systems and tasks.

\paragraph{\textbf{Lack of Unified Benchmarks}}

The lack of a common, user-friendly benchmark for evaluating different methods is another major challenge in RM. Researchers often assess their approaches within their own setups, which makes it difficult to conduct comprehensive and consistent comparisons across techniques. While some benchmarks do exist, they are either difficult to deploy on headless servers or are limited in the range of tasks they cover. This absence of standardized benchmarks impedes progress, as there is no reliable way to measure improvements or draw definitive conclusions across various studies. Additionally, existing datasets vary significantly in terms of observation modalities, action dimensions, and task settings, making them incompatible with one another.

\paragraph{\textbf{Dependence on Expert Data}}
Current robotic manipulation methods remain heavily reliant on expert data, such as labeled datasets and demonstrations. Often, systems require fine-tuning with domain-specific data before they can perform well on real-world tasks. Moreover, these systems typically lack sufficient adaptability during testing, with limited abilities for self-adaptation or continuous learning. This dependence on pre-collected data and the inability to continuously learn from new experiences restricts the scalability and flexibility of robotic manipulation techniques in real-world applications.

\paragraph{\textbf{Collaborative and Dexterous Manipulation}}
Dual-arm collaborative manipulation and high-degree-of-freedom manipulation are still in their early stages, representing an emerging area of research. These advancements are essential for enabling robots to coordinate and perform tasks with natural dexterity in real-life scenarios, as well as for the progression of humanoid robots.

\subsection{Conclusion}

In this survey, we provide a systematic and comprehensive review of RMP methodologies, highlighting key techniques, application scopes, learning objectives, data types, targeted challenges, strengths, and limitations. We analyzed 120 research papers to extract detailed data, focusing on: 1) the challenges addressed, 2) the core priors or tools, strengths, and limitations, 3) the type of input data and pretraining strategies, and 4) potential future research directions. The papers were categorized in a hierarchical manner based on their techniques. While significant progress has been made in RMP in recent years, we also identify ongoing challenges and open questions that set the stage for future investigation in this field.

This survey offers a comprehensive overview and valuable research resources for scholars in the field of RMP. However, there are some limitations of our review. Specifically, our focus was on RMP, and methods based on reinforcement learning were not covered. Additionally, since grasping is just one aspect of manipulation, we have excluded literature that focuses solely on grasp pose generation. Consequently, several valuable papers were not included. Despite our efforts to apply a systematic literature review methodology and conduct manual searches to ensure thorough inclusion, it is possible that some relevant papers were overlooked during the initial selection due to the vast volume of available literature.

\section{Acknowledgements}
This work was in part supported by the French Research Agency ANR, l’Agence Nationale de Recherche, through the projects Chiron (ANR-20-IADJ-0001-01), Aristotle (ANR-21-FAI1-0009-01), and Astérix (ANR-23-EDIA-0002), the French national investment prioritary program through the PSPC FAIR WASTE project, and the Franco-Chinese Research Center for Carbon Neutrality (Le Centre de Neutralité Carbone franco-chinois-CNC) through collaborative Artemis project. 


\def\UrlBreaks{\do\/\do-}
{\footnotesize
\bibliographystyle{IEEEtran}
\bibliography{ref}
}

\vspace{-5mm}
\begin{IEEEbiography}[{\includegraphics[width=0.90in,height=1.30in,clip,keepaspectratio]{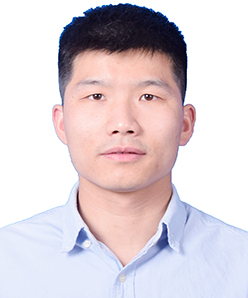}}]{Zezeng Li}
	received a B.S. degree from Beijing University of Technology (BJUT) in 2015 and a Ph.D. degree from Dalian University of Technology (DUT) in 2024. He is currently a postdoctoral fellow at the Ecole Centrale de Lyon (ECL).  His research interests include generative models and robotic manipulation.
\end{IEEEbiography}

\vspace{-3mm}
\begin{IEEEbiography}[{\includegraphics[width=0.90in,height=1.30in,clip,keepaspectratio]{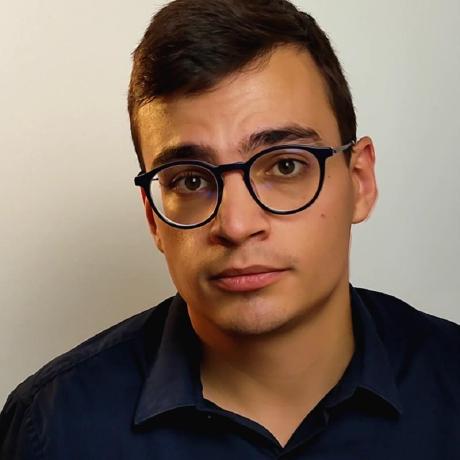}}]{Alexandre Chapin}
	received a M.S. degree from Institut National des Sciences Appliquées de Rennes (Insa Rennes) in 2022. He is currently pursuing a PhD at the Ecole Centrale de Lyon (ECL).  His research interests include structured image representation, generalization and robotic manipulation.
\end{IEEEbiography}
\vspace{-3mm}
\begin{IEEEbiography}[{\includegraphics[width=0.90in,height=1.30in,clip,keepaspectratio]{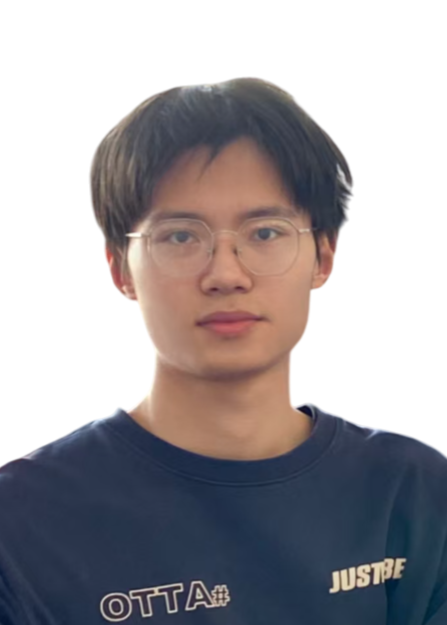}}]{Enda Xiang}
	received a B.S. degree from Beihang University (BUAA) in 2024. He is currently pursuing a Ph.D. degree at School of Computer Science and Engineering, Beihang University.  His research interests include imitation learning, grasping and robotic manipulation.
\end{IEEEbiography}
\vspace{-3mm}
\begin{IEEEbiography}[{\includegraphics[width=0.90in,height=1.30in,clip,keepaspectratio]{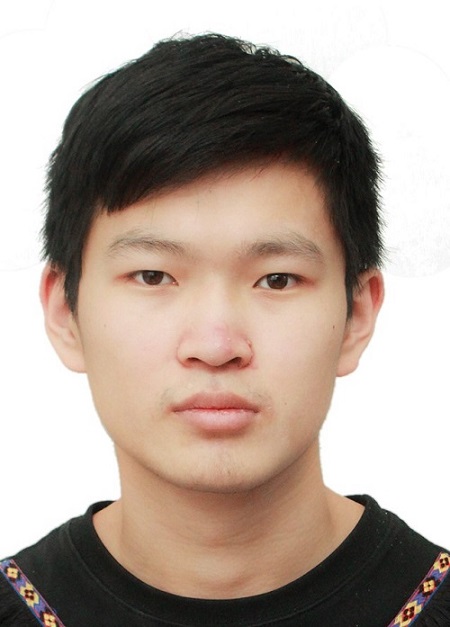}}]
    {Rui Yang} received a B.S. degree from Wuhan University in 2017 and an Engineering degree from École Centrale de Lyon (ECL) in 2020. He is currently pursuing a Ph.D. degree at LIRIS, École Centrale de Lyon. His research interests include continual learning and robotic manipulation.
\end{IEEEbiography}

\begin{IEEEbiography}[{\includegraphics[width=0.90in,height=1.30in,clip,keepaspectratio]{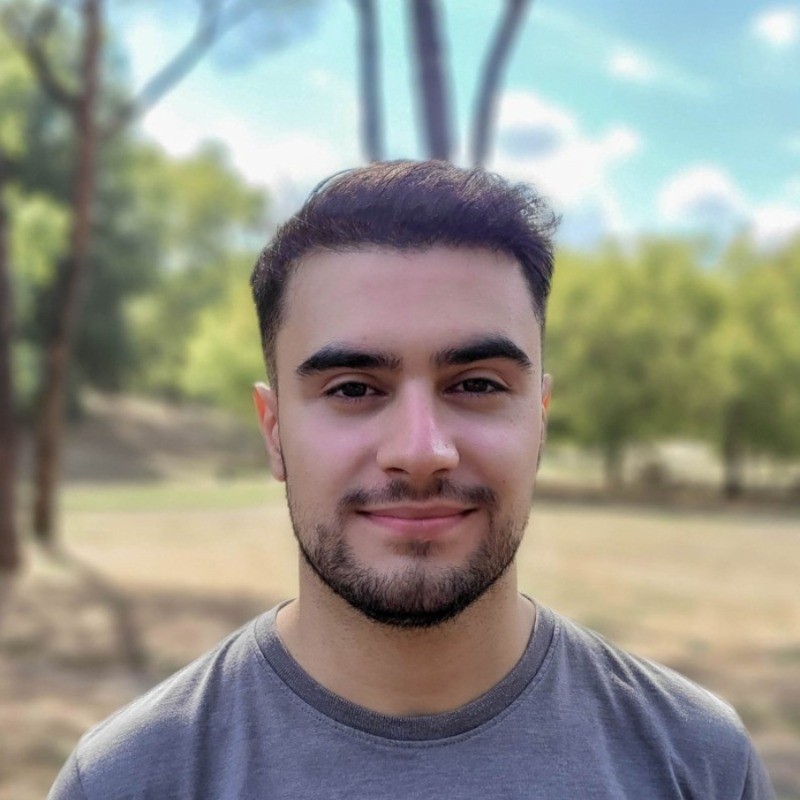}}]
    {Bruno Machado} received an M.S. degree from the École Nationale Supérieure de l'Électronique et ses Applications (ENSEA) in 2023. He is currently pursuing a Ph.D. at the École Centrale de Lyon (ECL).  His research is focused on self-supervised learning for robotic manipulation.
\end{IEEEbiography}

\begin{IEEEbiography}[{\includegraphics[width=1in,height=1.25in,clip,keepaspectratio]{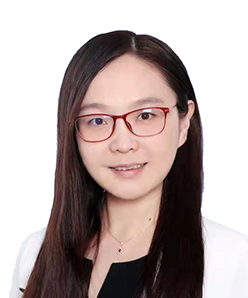}}]{Na Lei}
	received her B.S. degree in 1998 and a Ph.D. degree in 2002 from Jilin University. Currently, she is a professor at School of Software, Dalian University of Technology. Her research interest is the application of modern differential geometry and algebraic geometry to solve problems in engineering and medical fields. She mainly focuses on computational conformal geometry, computer mathematics, and its applications in computer vision, geometric modeling, and embodied intelligence.
\end{IEEEbiography}

\begin{IEEEbiography}[{\includegraphics[width=0.9in,height=1.2in,clip,keepaspectratio]{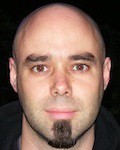}}]{Emmanuel Dellandrea}
	is Associate Professor at Ecole Centrale de Lyon, France, since 2004. He was awarded an M.S. and Engineering degrees in Computer Science from the Université de Tours, France, in 2000 followed by a Ph.D. in Computer Science in 2003. He obtained the Habilitation to Drive Research in 2020 from Université Lyon 1, France. His research at the LIRIS laboratory focuses on computer vision, and machine learning, with diverse applications including robotic manipulation.
\end{IEEEbiography}

\begin{IEEEbiography}[{\includegraphics[width=0.90in,height=1.50in,clip,keepaspectratio]{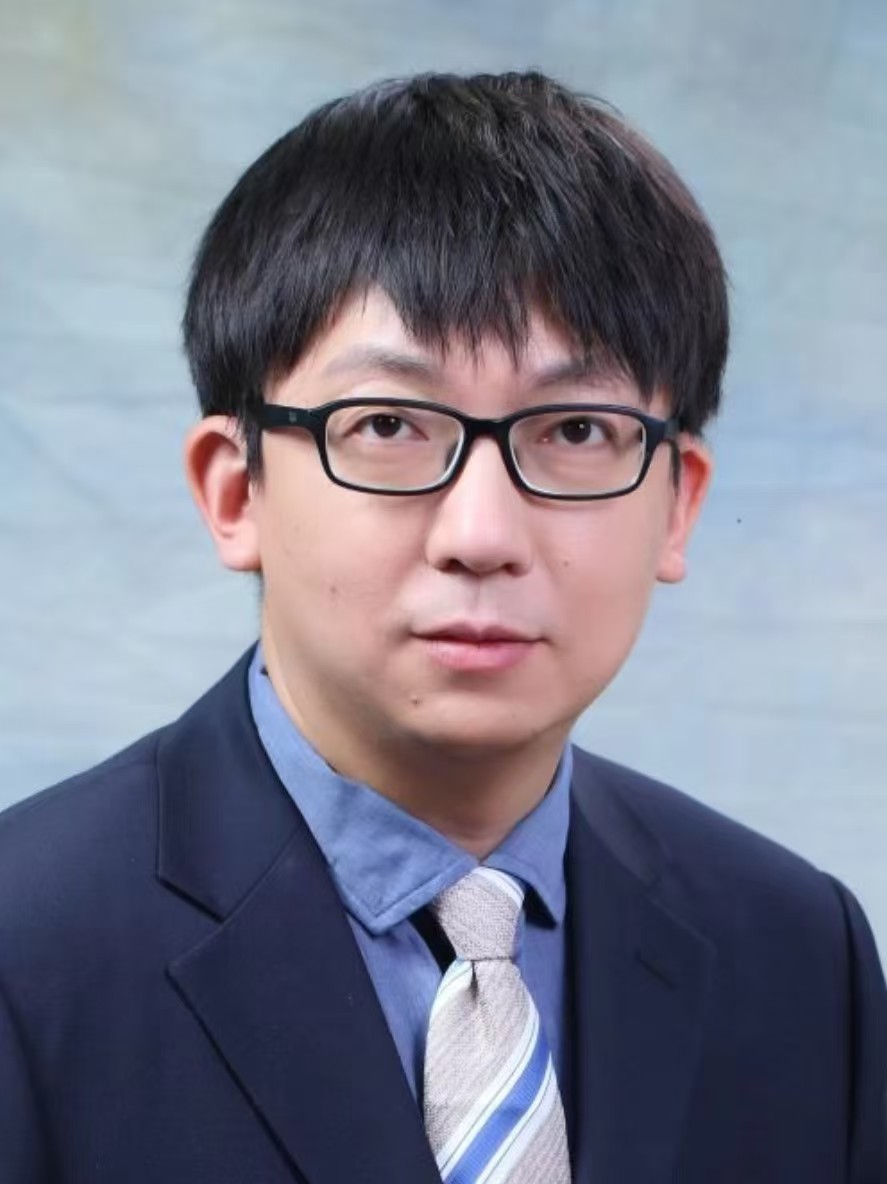}}]{Di Huang}
	received the B.S. and M.S. degrees in computer science from Beihang University, Beijing, China, in 2005 and 2008, respectively, and the Ph.D. degree in computer science from the École Centrale de Lyon, France, in 2011. He joined School of Computer Science and Engineering, Beihang University, as a Faculty Member, where he is currently a Professor. His research interests include computer vision, pattern recognition, and representation learning. He is a Senior Member of the IEEE.
\end{IEEEbiography}

\begin{IEEEbiography}[{\includegraphics[width=0.93in,height=1.50in,clip,keepaspectratio]{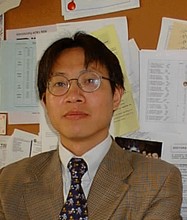}}]{Liming Chen}
	was awarded his B.Sc. degree in joint mathematics-computer science from the University of Nantes, France, in 1984, and his M.S. and Ph.D. degrees from the University of Paris 6, France, in 1986 and 1989. He first served as an Associate Professor with the Universite de Technologie de Compi`egne, before joining the Ecole Centrale de Lyon as a Professor in 1998, where he leads an Advanced Research Team in multimedia computing and pattern recognition. His current research interests include computer vision and multimedia, and in particular face analysis, image and video categorization, affective computing, and robotic manipulation. He is a Senior Member of the IEEE.
\end{IEEEbiography}

\end{document}